\documentclass[11pt]{article}

\usepackage[final]{acl}

\usepackage{times}
\usepackage{latexsym}
\usepackage{enumitem}

\usepackage[T1]{fontenc}

\usepackage[utf8]{inputenc}

\usepackage{microtype}

\usepackage{inconsolata}

\usepackage{graphicx}
\usepackage{amsmath}
\usepackage{multirow}
\usepackage{makecell}
\usepackage{subcaption}
\usepackage{caption}
\usepackage{kotex}
\usepackage{colortbl}
\usepackage{algorithm}        
\usepackage{algpseudocode}
\usepackage{booktabs}
\usepackage{amssymb}

\usepackage{xcolor}
\usepackage{pifont}

\usepackage{tabularx}

\newcommand{\cmark}{\textcolor{green!60!black}{\ding{51}}}
\newcommand{\xmark}{\textcolor{red!70!black}{\ding{55}}}

\usepackage[table]{xcolor}
\definecolor{topaz}{RGB}{255,200,124}
\definecolor{flax}{RGB}{238,220,130}

\usepackage[colorinlistoftodos,prependcaption]{todonotes} 
\presetkeys{todonotes}{inline, color=topaz!40, bordercolor=topaz}{}

\usepackage{tikz}
\newcommand*\circled[1]{\tikz[baseline=(char.base)]{
            \node[shape=circle,draw,inner sep=0.5pt] (char) {#1};}}

\usepackage{dblfloatfix}

\title{Robust Multimodal Sentiment Analysis with Incomplete Modalities via Semantic-aware Completeness based Reconstruction}

\author{
  Han-Jun Choi\thanks{Equal contribution.}
  \and Byunggill Joe\footnotemark[1]
  \and Saim Shin
  \and Jin Yea Jang\thanks{Corresponding author.} \\
  Korea Electronics Technology Institute (KETI) \\
  Seongnam, South Korea \\
  \texttt{\{hanjun\_c, byunggill, sishin, jinyea.jang\}@keti.re.kr}
}

\begin{document}
\maketitle

\begin{abstract}
Recent multimodal sentiment analysis studies increasingly adopt text-centric fusion approaches to exploit the rich sentiment information inherent in the textual modality. However, these approaches often suffer from performance degradation during inference due to partially missing or noisy data in real-world scenarios, especially when sentiment-related cues are missing. To address this issue, we introduce a new completeness estimation approach that quantifies the degree of sentiment-relevant information preserved in incomplete data to guide the reconstruction of missing semantics. Furthermore, we propose a training strategy that stabilizes multi-task learning while jointly optimizing sentiment prediction and completeness estimation. Extensive experiments and in-depth analyses on three benchmark datasets demonstrate that the proposed approach enables more accurate semantic reconstruction, leading to more precise sentiment prediction.
\end{abstract}    
\section{Introduction}
\label{sec:intro}

\begin{figure}[t!]
    \centering
    \includegraphics[width=1.0\linewidth]{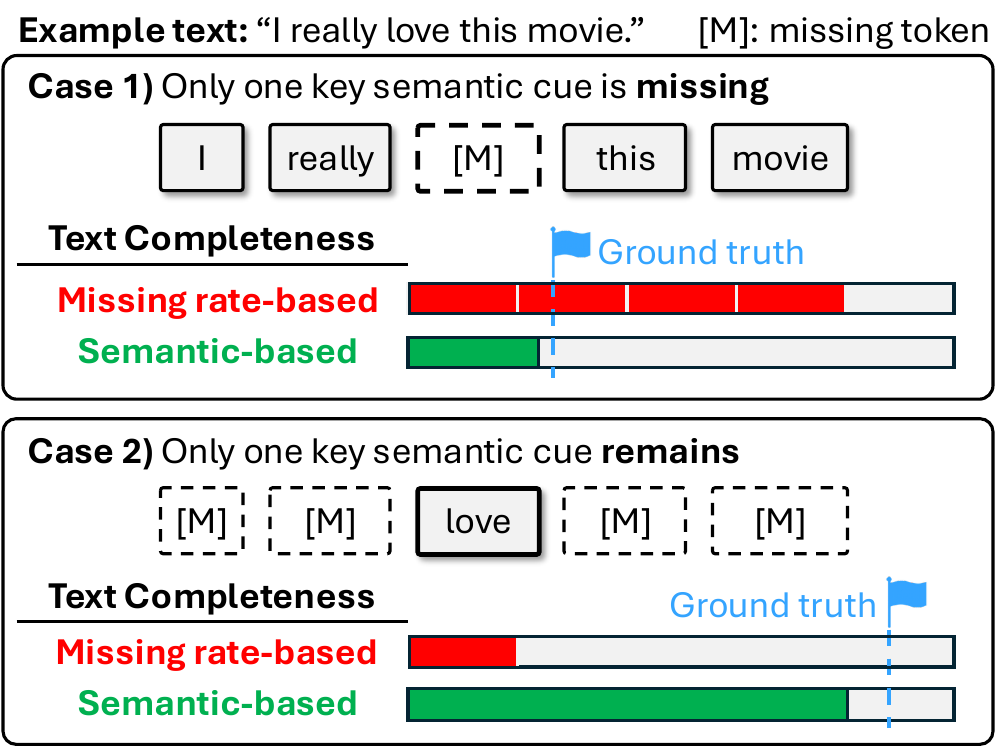}
    \caption{Comparison between missing rate–based and semantic–based completeness. The former is computed as $1 - r$, where $r \in [0, 1)$ denotes the missing rate, while the latter is estimated by our proposed method.}
    \label{fig:motivation}
\end{figure}

Building on the rich sentiment-relevant information in text~\cite{Hazarika_2022_NAACL, wei2023tackling}, recent Multimodal Sentiment Analysis (MSA) studies have increasingly explored text-centric approaches~\cite{han2021bibimodal, wang2023tetfn, 9797846, zhang2023language} that treat text as the dominant modality while considering audio and vision modalities as auxiliary sources. Although these approaches have shown strong performance with complete data, they suffer substantial performance degradation when modalities are partially missing or noisy, particularly when the textual modality is incomplete.

To address this issue, several reconstruction-based approaches have been proposed~\cite{yuan2021transformer, zhang2024robust, zhu-etal-2025-proxy}. In particular, ~\citet{zhang2024robust} proposed a missing rate-based completeness estimation method to utilize the missing rate as a reconstruction weight, assuming that a higher missing rate indicates lower completeness, and vice versa. However, we argue that relying solely on the missing rate may fail to accurately capture semantic informativeness. 

Figure~\ref{fig:motivation} highlights this limitation through two contrasting cases. In Case 1, a missing key sentiment cue [\textit{`love'}] leads to substantial semantic loss despite a low missing rate, causing an overestimation of missing rate-based completeness. On the other hand, in Case 2, it is underestimated even though the sentiment meaning remains intact. Such inaccurate completeness may lead to unreliable reconstruction, which highlights the necessity of a semantic-aware completeness approach that determines completeness based on whether key sentiment cues are missing.

Motivated by the above observations, we propose a semantic-aware completeness estimation approach for more reliable reconstruction of missing textual semantics. Specifically, we first estimate the completeness of incomplete text through a semantic completeness estimator. To supervise the estimator, we introduce Target Probability-based Semantic Completeness (TPSC), which is a pseudo-labeling strategy that generates completeness labels. Meanwhile, the Importance-aware Proxy Feature Generator (IPFG) generates proxy features from the auxiliary modalities by adaptively weighting their contributions for text reconstruction. The estimated completeness is then used as a weighting factor to adaptively combine proxy features and incomplete text, producing reconstructed textual representations. Furthermore, we propose an Alternating Optimization Strategy (AOS) to mitigate gradient conflicts that stem from jointly optimizing the sentiment prediction and completeness estimation tasks on a shared encoder. The contributions of this paper are summarized as follows:

\begin{itemize}
\item To the best of our knowledge, we are the first to introduce a semantic-aware completeness estimation approach that quantifies the degree of semantic preservation in incomplete data.

\item We propose an optimization strategy that mitigates gradient conflicts in hierarchical multi-task learning, thereby stabilizing training between the completeness estimation and sentiment prediction tasks.

\item Extensive experiments on three benchmark datasets with varying missing rates show that our proposed method consistently outperforms 12 competitive baselines, demonstrating the effectiveness of the proposed completeness label in real-world data scenarios.
\end{itemize}
\section{Related Work}
\label{sec:related_work}
Recent studies in MSA have explored various multimodal fusion strategies to integrate textual, acoustic, and visual information~\cite{zadeh-etal-2017-tensor, liu-etal-2018-efficient-low, tsai-etal-2019-multimodal, 10.1145/3394171.3413678, yu2021learning, sun2022cubemlp}. These approaches aim to capture cross-modal interactions and leverage complementary information across modalities to improve sentiment prediction. Another line of work focuses on text-centric approaches~\cite{han2021bibimodal, wang2023tetfn, 9797846, zhang2023language}, which treat text as the dominant modality and utilize audio and visual signals as auxiliary sources to enhance textual representations. However, most existing methods assume that all modalities are fully available during inference, which limits their robustness in real-world scenarios. This issue becomes particularly critical for text-centric approaches, as the loss of important semantic information in text directly undermines the effectiveness of text-centric fusion.

To mitigate the above issue, recent studies have explored reconstruction-based approaches that aim to restore semantic information from missing modalities~\cite{yuan2021transformer, lin-hu-2023-missmodal, zhang2024robust, zhu-etal-2025-proxy, yang2025decoupling, shi2025textguided, li-etal-2025-tf, lin2025cyin}. In particular, LNLN~\cite{zhang2024robust} introduces a reconstruction method that restores missing textual semantics using proxy features derived from audio and vision inputs, where the reconstruction weight is estimated from the missing rate of the incomplete text. P-RMF~\cite{zhu-etal-2025-proxy} introduces a proxy-driven strategy that dynamically reconstructs incomplete multimodal inputs through cross-modal feature generation and adaptive fusion, further improving robustness under uncertain missing conditions. Additionally, TF-Mamba~\cite{li-etal-2025-tf} proposes a text-enhanced Mamba-based framework that reconstructs missing textual semantics by aligning and enhancing auxiliary modalities through a text-aware modality enhancement module while modeling multimodal dependencies via efficient Mamba-based fusion. While these methods improve robustness to missing modalities, they do not explicitly account for the semantic information loss in incomplete text.
\begin{figure*}[t!]
    \centering
    \includegraphics[width=1.0\textwidth]{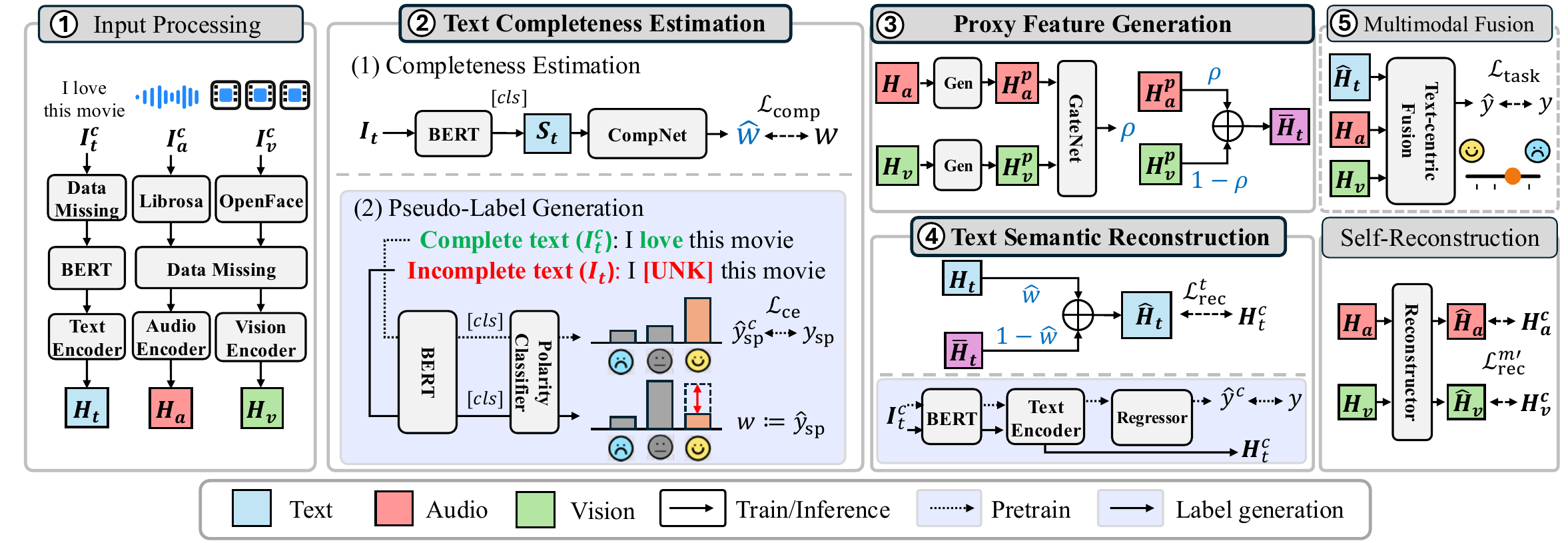}
    \caption{Overview of the proposed framework. Models in the light purple boxes are used only during pretraining or pseudo-labeling and are not part of the TCMR training pipeline.}
    \label{fig:overview}
\end{figure*}

\section{Proposed Method}

\label{sec:method}

\subsection{Problem Definition}
Given multimodal inputs from text (t), audio (a), and vision (v) derived from the same utterance, the MSA task aims to predict a continuous sentiment score $y\in \mathbb{R}$.

\subsection{Framework Overview}
In this work, we introduce a \textbf{T}ext \textbf{C}ompleteness-based \textbf{M}issing \textbf{R}econstruction (\textbf{TCMR}) framework that estimates textual completeness from incomplete text and reconstructs the missing semantic information by adaptively weighting proxy and incomplete text features based on the estimated completeness.

The overall workflow of TCMR is illustrated in \autoref{fig:overview}. First, each modality input is randomly masked and then embedded into feature representations via modality-specific encoders (\circled{1}). 
Next, a completeness estimator predicts how much semantic information remains in incomplete text and reconstructs missing semantics using proxy features derived from auxiliary modalities, weighted by the predicted completeness (\circled{2}–\circled{4}). Lastly, a multimodal fusion module performs text-centric fusion to produce the final sentiment score (\circled{5}). Note that the multimodal fusion module is adopted from LNLN, as designing a new fusion module is not the main objective of this work (see Appendix~\ref{sec:fusion}).

\subsection{Input Processing}
Raw multimodal inputs are processed into high-level embeddings (Figure~\ref{fig:overview}~\circled{1}). First, each complete multimodal input $I_m^c$ for modality $m \in \{v, a, t\}$, is processed by modality-specific feature extractors: BERT for text, Librosa for audio, and OpenFace for vision~\cite{devlin2019bert, mcfee2015librosa, baltrusaitis2016openface}.

Following prior works~\cite{zhang2024robust, zhu-etal-2025-proxy}, we adopt a random missing scenario to generate incomplete features $X_m \in \mathbb{R}^{T_m \times d_m}$, where $T_m$ and $d_m$ denote the sequence length and feature dimension. For text, a proportion of tokens in $I_t^c$ are randomly replaced with \texttt{[UNK]} according to a missing rate sampled from $[0,1)$. For audio and vision features, a proportion of the sequence is replaced with zero vectors. Subsequently, the incomplete features are processed by modality-specific Transformer encoders~\cite{vaswani2017attention}  to obtain the high-level embeddings $H_m \in \mathbb{R}^{T \times d}$:
\begin{equation}
    H_m = \text{Transformer}(X_m; \theta^{\text{trans}}_m)
    \label{eq:high_level_emb}
\end{equation}

\subsection{Missing Semantic Reconstruction}
\label{sec:completeness_reconstruction}
In this section, we elaborate on the completeness–guided semantic reconstruction pipeline of TCMR (Figure~\ref{fig:overview}~\circled{2}–\circled{4}), which is the core contribution of our work.
\paragraph{Text Completeness Estimation}
To estimate the completeness of incomplete text, we design a semantic completeness estimator \texttt{CompNet}, which consists of fully connected layers followed by a sigmoid activation (Figure~\ref{fig:overview}~\circled{2} --1). 

Given the BERT \texttt{[CLS]} embedding $S_t \in \mathbb{R}^{d}$, \texttt{CompNet} estimates a completeness weight $\hat{w}\in[0,1]$:

\begin{equation} \hat w = \text{CompNet}(S_t; \theta^{\mathrm{comp}}), 
\label{eq:compnet}
\end{equation}
The model is trained to minimize mean squared error between the predicted score $\hat{w}$ and the completeness label $w$:
\begin{equation}
\mathcal{L}_{\text{comp}} = \frac{1}{N} \sum_{i=1}^{N} \left\| \hat w^{(i)} - w^{(i)} \right\|^2,
\label{eq:tce_loss}
\end{equation}
where $i$ denotes the data sample index.

\paragraph{Pseudo-Label Generation}
\label{sec:label_gen}
In our study, we define textual completeness as the extent to which sentiment-related semantic information is preserved in incomplete text. Following this definition, we introduce a pseudo-labeling strategy to generate completeness labels $w$ for training \texttt{CompNet} (Figure~\ref{fig:overview}~\circled{2} --2).

To quantify the completeness, we extend the concept of True Class Probability (TCP)~\cite{corbiere2019confidence}, defined as the probability assigned to the target class. Our hypothesis is that if a classifier trained on complete text assigns a high TCP to a given complete text, the TCP for the corresponding incomplete text will remain high as long as sentiment-relevant tokens are preserved. Conversely, the TCP may substantially decrease if key sentiment tokens are missing.

Based on this insight, we design a sentiment polarity classifier that takes only the text modality as input. First, the classifier is trained on the complete text $I_t^c$:
\begin{align}
\hat{y}_{\text{sp}} &= \text{Classifier}(\text{BERT}^{\text{cls}}(I_t^c); \theta^{\text{cls}})\\
\mathcal{L}_{\text{ce}} &= - \sum_{c=1}^{C} y_{sp}^{(c)} \log \hat{y}_{sp}^{(c)}
\end{align}
where $y_{sp}$ denotes the polarity class obtained by discretizing the continuous sentiment score according to predefined thresholds. After training, we perform inference on incomplete text using the pretrained classifier, and the probability assigned to the ground-truth polarity class $y_{sp}$ is used as the pseudo completeness label $w \in [0,1]$:
\begin{equation}
w^{(i)} \triangleq p\!\left(y^{(i)}_{\text{sp}}\mid \text{BERT}^{\text{cls}}(I_t);\theta^{\text{cls}}_{\text{pre}}\right),
\end{equation}
where $ \theta^{\text{cls}}_{\text{pre}}$ are the pretrained classifier parameters. The definition of sentiment polarity classes and the choice of the optimal number of classes are provided in Appendix~\ref{sec:definition_of_sp} and Appendix~\ref{TPSC_cls}, respectively.

\paragraph{Proxy Feature Generation}
Prior work~\cite{zhang2024robust} generates proxy features without considering the relative importance of auxiliary modalities for reconstruction. However, the contributions of each modality to the reconstruction can vary under random missing conditions.

Therefore, we design an Importance-aware Proxy Feature Generator (IPFG), which consists of Transformer-based modality-specific generators and an MLP-based gating network with a sigmoid:
\begin{align}
H_{m'}^{p} &= \mathrm{Gen}(H_{m'};\, \theta_{m'}^{\mathrm{gen}}),\\
\rho &= \mathrm{Gate}([H_{a}^{p},H^p_v];\, \theta^{\mathrm{gate}}), \\
\bar H_t &= \rho \odot H_a^{p} + (1-\rho)\odot H_v^{p}.
\end{align}
where $m' \in \{a,v\}$ denotes the auxiliary modalities, $\odot$ denotes element-wise multiplication, and $\rho \in [0,1]$ is a gating weight that adaptively balances the contributions of the auxiliary modalities.

\paragraph{Semantic Reconstruction}
The reconstructed text representation $\hat{H}_t \in \mathbb{R}^{T \times d}$ is generated by integrating the incomplete text representation $H_t$ with the generated proxy feature $\bar H_t$, guided by the predicted completeness weight $\hat{w}$ from \texttt{CompNet}:
\begin{equation}
    \hat{H}_t = \hat w  H_t + (1 - \hat w)\bar{H}_t.
\end{equation}
To encourage the reconstructed features to be similar to the corresponding complete features, we minimize the mean squared error loss $\mathcal{L}^t_{rec}$. For the auxiliary modalities, following prior work~\cite{zhang2024robust}, we adopt a self-reconstruction method that takes $H_{m'}$ as input and predicts the corresponding complete representation $\hat{H}_{m'}$. The reconstruction losses are defined as:
\begin{equation}
\mathcal{L}_{\text{rec}}^{m}
=
\frac{1}{N}
\sum_{i=1}^{N}
\left\| \hat{H}_{m}^{(i)} - H_{m}^{c,(i)} \right\|^{2},
\end{equation}
where $H_m^{c,(i)}$ denotes the complete feature of modality $m$ obtained from a model trained with the corresponding complete input.

\subsection{Training Objective}
TCMR predicts the final sentiment score $\hat{y}$ through the text-centric multimodal fusion module. 
The model is trained using the following loss:
\begin{equation}
\mathcal{L}_{\text{task}} =
\frac{1}{N}\sum_{i=1}^{N}
\left\|
\hat{y}^{(i)} - y^{(i)}
\right\|^2 .
\label{eq:task_loss}
\end{equation}
The overall training objective is defined as:
\begin{equation}
\mathcal{L}_{\text{total}} =
\alpha\mathcal{L}_{\text{comp}}
+\beta\mathcal{L}^{t}_{\text{rec}}
+\gamma\mathcal{L}_{\text{rec}}^{m'}
+\sigma\mathcal{L}_{\text{task}} .
\label{eq:overall_loss}
\end{equation}
where $\alpha$, $\beta$, $\gamma$, and $\sigma$ are hyperparameters.

\begin{algorithm}[t]
\caption{Alternating Optimization Strategy}
\label{alg:aos-compact}
\textbf{Input:} dataset $\mathcal{D}$, epochs $E$, batch size $B$, weights $\alpha,\beta,\gamma,\sigma$.\\
\textbf{Params:} $\Theta^{\text{comp}}=\theta^{\text{comp}}\cup\theta^{\text{bert}}$, $\Theta^{\text{other}}$ for the remaining modules including $\theta^{\text{bert}}$.\\
\textbf{Opt:} $\mathrm{Opt}_{\text{comp}}$, $\mathrm{Opt}_{\text{other}}$.
\begin{algorithmic}[1]
\For{$e=1$ to $E$}
\Statex \textbf{Phase 1:} optimize $\Theta^{\text{comp}}$ to minimize $\mathcal{L}_{\text{comp}}$
\For{mini-batch $\mathcal{B}\subset\mathcal{D}$, $|\mathcal{B}|=B$}
    \State $\mathcal{L}_{\text{comp}} \gets \mathrm{LossComp}(\mathcal{B};\,\Theta^{\text{comp}})$
    \State $\mathcal{L}_{\text{comp}} \gets \alpha\,\mathcal{L}_{\text{comp}}$
    \State $\mathrm{Opt}_{\text{comp}}.\mathrm{step}\!\left(\nabla_{\Theta^{\text{comp}}}\mathcal{L}_{\text{comp}}\right)$
\EndFor
\Statex \textbf{Phase 2:} optimize $\Theta^{\text{other}}$ to minimize $\mathcal{L}_{\text{other}}$
\For{mini-batch $\mathcal{B}\subset\mathcal{D}$, $|\mathcal{B}|=B$}
    \State $\mathcal{L}_{\text{task}} \gets \mathrm{LossTask}(\mathcal{B};\,\Theta^{\text{other}})$
    \State $\mathcal{L}_{\text{rec}}^{t} \gets \mathrm{LossRecText}(\mathcal{B};\,\Theta^{\text{other}})$
    \State $\mathcal{L}_{\text{rec}}^{m'} \gets \mathrm{LossRecAux}(\mathcal{B};\,\Theta^{\text{other}})$
    \State $\mathcal{L}_{\text{other}} \gets \beta\,\mathcal{L}_{\text{rec}}^{t} + \gamma\,\mathcal{L}_{\text{rec}}^{m'} + \sigma\,\mathcal{L}_{\text{task}}$
    \State $\mathrm{Opt}_{\text{other}}.\mathrm{step}\!\left(\nabla_{\Theta^{\text{other}}}\mathcal{L}_{\text{other}}\right)$
\EndFor
\EndFor
\end{algorithmic}
\end{algorithm}

\subsection{Alternating Optimization Strategy}
\label{sec:aos}
We found that naively minimizing $\mathcal{L}_{\text{total}}$ in an end-to-end manner often leads to unstable training, where $\mathcal{L}_{\text{comp}}$ is effectively ignored and \texttt{CompNet} remains stuck at a poor local minimum. We attribute this behavior to gradient conflict~\cite{yu2020gradient}. 
This phenomenon commonly arises in multi-task learning, where the gradients of a dominant task suppress the optimization of relatively weaker tasks. 
In our case, $\mathcal{L}_{\text{task}}$ can also be minimized through shortcut solutions that bypass learning completeness estimation, causing the optimization of $\mathcal{L}_{\text{comp}}$ to be dominated during training.

To address this issue, we introduce a simple yet effective training strategy called Alternating Optimization Strategy (AOS). We first partition the total objective $\mathcal{L}_{\text{total}}$ into two groups according to their respective roles: \begin{equation} \mathcal{L}_{\text{comp}}, \quad \mathcal{L}_{\text{other}} = \beta\,\mathcal{L}_{\text{rec}}^{t} + \gamma\,\mathcal{L}_{\text{rec}}^{m'} + \sigma\,\mathcal{L}_{\text{task}} . \end{equation} 

\begin{figure}[t]
    \centering
    \includegraphics[width=\linewidth]{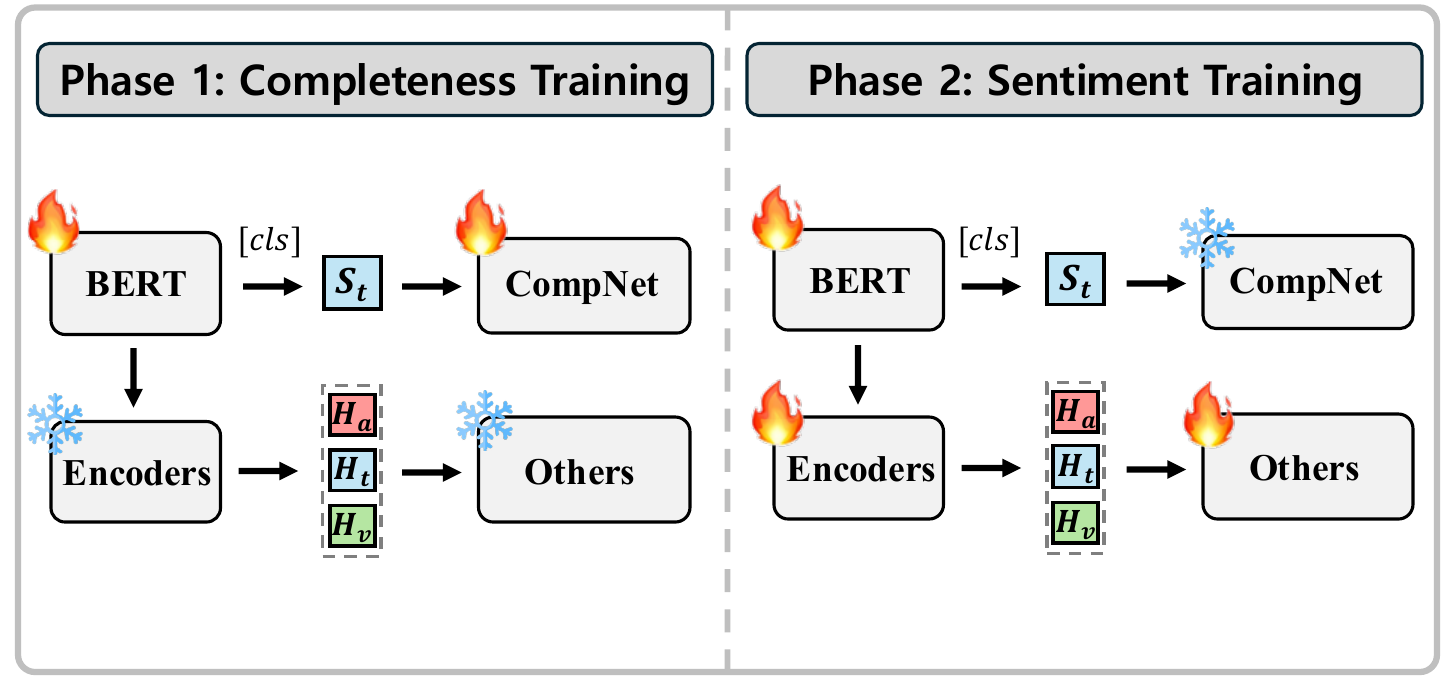}
    \caption{Illustration of the AOS optimization process.}
    \label{fig:aos_vis}
\end{figure}

As shown in Algorithm~\ref{alg:aos-compact}, AOS performs a two-phase optimization within each training epoch. In Phase 1, we focus on completeness learning by minimizing $\mathcal{L}_{\text{comp}}$, while fixing the remaining parameters and optimizing only $\Theta^{\text{comp}}$. In Phase 2, we fix the learned $\theta^{\text{comp}}$ and optimize $\Theta^{\text{other}}$ by minimizing $\mathcal{L}_{\text{other}}$. At this stage, the model focuses on reconstruction and sentiment prediction based on the completeness estimation learned in Phase 1. By alternately optimizing the two objectives, AOS stabilizes the training of \texttt{CompNet} and enables the remaining modules to be effectively trained according to their respective objectives. Figure~\ref{fig:aos_vis} illustrates this alternating optimization process, and further experimental analysis of gradient conflicts is provided in Appendix~\ref{sec:aos_appendix}.
\section{Experiments}
\label{sec:experiments}

\subsection{Experimental Setup}

\paragraph{Dataset and Evaluation Metrics}

We conduct experiments on three MSA benchmark datasets: MOSI~\cite{zadeh2016multimodal}, MOSEI~\cite{bagher-zadeh-etal-2018-multimodal}, and SIMS~\cite{yu2020chsims}. 
In MOSI and MOSEI, sentiment scores are annotated with continuous values ranging from $-3$ (strongly negative) to $+3$ (strongly positive), while SIMS provides sentiment labels on a scale from $-1$ (negative) to $+1$ (positive). Model performance is evaluated using Acc-2, Acc-5, Acc-7, F1, MAE, and Corr on MOSI and MOSEI, and Acc-2, Acc-3, Acc-5, F1, MAE, and Corr on SIMS. Dataset statistics and metric definitions are provided in Appendix~\ref{sec:data_stas} and Appendix~\ref{sec:eval_metrics}, respectively.

\paragraph{Training and Evaluation Settings}
Following previous studies~\cite{zhang2024robust, zhu-etal-2025-proxy}, we adopt a \textit{partial random missing} scenario. During training, each modality is randomly erased according to a missing rate $r$ sampled from a uniform distribution over the range $[0, 1.0)$. For each of the three seeds, the best model is selected based on its performance at a missing rate of $r = 0.5$ on the validation set. For testing, we conduct experiments across missing rates ranging from $0$ to $0.9$ with an increment of $0.1$, resulting in ten evaluation settings for each seed.

\begin{table*}[ht!]
\centering
\setlength{\tabcolsep}{2.5pt}
\caption{Comparison of the overall performance on MOSI and MOSEI datasets.}
\label{tab:main_results}
\resizebox{\textwidth}{!}{%
\begin{tabular}{l|cccccc|cccccc}
\toprule
\multirow{2}{*}{Method} & \multicolumn{6}{c|}{\textbf{MOSI}} & \multicolumn{6}{c}{\textbf{MOSEI}} \\
 & Acc-7 & Acc-5 & Non0 Acc / F1 & Has0 Acc / F1 & MAE & Corr & Acc-7 & Acc-5 & Non0 Acc / F1 & Has0 Acc / F1 & MAE & Corr \\
\midrule
MISA     & 29.03 & 31.61 & 68.77 / 68.67 & 67.94 / 67.72 & 1.1637 & 47.57 & 43.89 & 44.43 & 72.22 / 68.21 & 74.16 / 71.24 & 0.7346 & 43.57 \\
Self-MM  & 30.38 & 33.69 & 68.83 / 68.63 & 68.65 / 69.34 & 1.1843 & 47.25 & 46.45 & 47.36 & 73.02 / 71.43 & 72.66 / 71.82 & 0.6818 & 53.68 \\
MMIM     & 30.67 & 34.31 & 70.19 / 69.87 & 69.59 / 69.15 & 1.1649 & 49.01 & 44.89 & 45.42 & 74.29 / 73.19 & 73.46 / 73.19 & 0.7032 & 52.75 \\
CENet    & 29.49 & 32.90 & 69.88 / 69.94 & 69.43 / 69.39 & 1.1801 & 48.24 & \textbf{47.36} & \textbf{48.24} & 77.01 / 77.26 & 75.96 / 76.23 & 0.6622 & 58.24 \\
TETFN    & 29.86 & 32.56 & 70.62 / 70.67 & 69.85 / 69.79 & 1.1327 & 49.16 & 46.78 & 47.83 & 77.87 / 77.45 & 76.11 / 76.37 & 0.6741 & 58.31 \\
TFR-Net  & 28.22 & 30.31 & 70.88 / 70.73 & 70.18 / 69.93 & 1.1454 & 50.05 & 46.08 & 46.47 & 75.46 / 74.10 & 74.18 / 73.61 & 0.6784 & 56.24 \\
ALMT     & 29.57 & 32.05 & 71.51 / 71.48 & 70.52 / 70.39 & 1.1671 & 47.57 & 46.66 & 47.37 & 76.83 / 76.41 & 74.47 / 74.84 & 0.6749 & 56.45 \\
\midrule
LNLN     & 31.36 & 34.43 & 70.00 / 69.99 & 69.51 / 69.41 & 1.1450 & 48.08 & 46.36 & 47.14 & 77.79 / 77.27 & 76.43 / 76.58 & 0.6698 & 58.17 \\
P-RMF    & 28.91 & 31.24 & 69.91 / 69.57 & 69.28 / 68.84 & 1.1302 & 50.07 & 47.01 & 47.82 & \textbf{78.07 / 77.76} & 77.18 / 76.91 & 0.6667 & 58.45 \\
TF-Mamba & 28.85 & 31.46 & 71.05 / 71.11 & 70.73 / 70.69 & 1.1228 & 51.97 & 45.11 & 46.09 & 76.11 / 76.17 & 72.54 / 73.49 & 0.6891 & 57.51 \\
\midrule
TCMR-MRSC & 30.39 & 33.36 & 70.84 / 70.95 & 69.81 / 69.82 & 1.1061 & 50.86 & 46.61 & 47.46 & 77.65 / 76.74 & 76.91 / 76.64 & 0.6673 & 58.46 \\
TCMR-LDSC & 31.63 & 35.38 & 70.61 / 70.55 & 69.83 / 69.58 & 1.1322 & 49.07 & 47.13 & 47.87 & 77.52 / 76.97 & 76.16 / 76.29 & 0.6641 & 58.51 \\
\textbf{TCMR-TPSC} & \textbf{32.80} & \textbf{36.73} & \textbf{72.01} / \textbf{71.67} & \textbf{70.93} / \textbf{70.49} & \textbf{1.0721} & \textbf{52.28} & 47.27 & 48.16 & 77.99 / 76.99 & \textbf{77.59} / \textbf{77.21} & \textbf{0.6614} & \textbf{58.63} \\
\bottomrule
\end{tabular}
}
\end{table*}

\subsection{Main Experimental Results}
\paragraph{Baseline Models}

We select baseline models with publicly available implementations to ensure reproducibility. For a fair comparison, we reproduce all baseline models using publicly available implementations. For MISA, Self-MM, MMIM, CENet, TETFN, TFR-Net, and ALMT, we use the MMSA implementations~\cite{Mao_2022_ACL}. For LNLN\footnote{\url{https://github.com/Haoyu-ha/LNLN}}, P-RMF\footnote{\url{https://github.com/aoqzhu/P-RMF}}, and TF-Mamba\footnote{\url{https://github.com/codemous/TF-Mamba}}, we use their official repositories. In addition, we implement three variants of TCMR with different completeness labeling strategies to analyze their effectiveness:

\begin{itemize}
    \item \textbf{TCMR-TPSC}: estimates completeness using \textit{Target Probability-based Semantic Completeness} (TPSC), as defined in Section~\ref{sec:completeness_reconstruction}. Note that unless otherwise specified, TCMR refers to TCMR-TPSC for simplicity.
    \item \textbf{TCMR-MRSC}: estimates completeness using \textit{Missing Rate-based Semantic Completeness} (MRSC), which measures completeness according to the missing rate of text.
    \item \textbf{TCMR-LDSC}: To better analyze the effectiveness of TPSC, we design \textit{Lexicon-Derived Semantic Completeness} (LDSC). LDSC generates completeness labels based on sentiment scores of individual tokens obtained from SentiWordNet~\cite{baccianella2010sentiwordnet}. Specifically, we first compute the sentiment strength of the complete text by accumulating the sentiment scores of all tokens. Then, the same procedure is applied to the incomplete text. As a result, the completeness label is defined as the ratio between the two strengths. Further implementation details are provided in Appendix~\ref{sec:ldsc}.
\end{itemize}

\begin{table}[t]
\centering
\small
\resizebox{\columnwidth}{!}{
\begin{tabular}{lcccccc}
\toprule
Method & Acc-5 & Acc-3 & Acc-2 & F1 & MAE & Corr \\
\midrule
MISA & 32.97 & 56.05 & 71.25 & 66.95 & 0.5731 & 0.319 \\
Self-MM & 33.57 & 55.82 & 69.02 & 68.38 & 0.5243 & 0.363 \\
MMIM & 30.33 & 55.04 & 69.46 & 66.74 & 0.5629 & 0.301 \\
CENet & 21.61 & 53.48 & 69.23 & 56.97 & 0.6304 & 0.014 \\
TETFN & \textbf{34.91} & 56.01 & 70.85 & 69.17 & 0.5439 & 0.342 \\
TFR-Net & 27.21 & 54.26 & 69.37 & 56.82 & 0.6258 & 0.091 \\
ALMT & 30.99 & 53.23 & 70.65 & 67.34 & 0.5445 & 0.311 \\
\midrule
LNLN & 32.17 & 56.77 & 71.51 & 69.21 & 0.5423 & 0.345 \\
P-RMF & 32.72 & 57.30 & 70.58 & \textbf{70.36} & 0.5432 & 0.361 \\
TF-Mamba & 33.05 & 55.18 & 69.88 & 69.24 & 0.5235 & 0.359 \\
\midrule
TCMR-MRSC & 33.27 & 55.46 & 70.03 & 68.76 & 0.5446 & 0.341 \\
TCMR-TPSC & 31.15 & \textbf{57.45} & \textbf{72.81} & 68.88 & \textbf{0.5212} & \textbf{0.379} \\
\bottomrule
\end{tabular}
}
\caption{Comparison of the overall performance on the SIMS dataset. Note: TCMR-LDSC is excluded since it is based on an English sentiment lexicon.}
\label{tab:sims_results}
\end{table}

\begin{figure}[t]
    \centering
    \includegraphics[width=\linewidth]{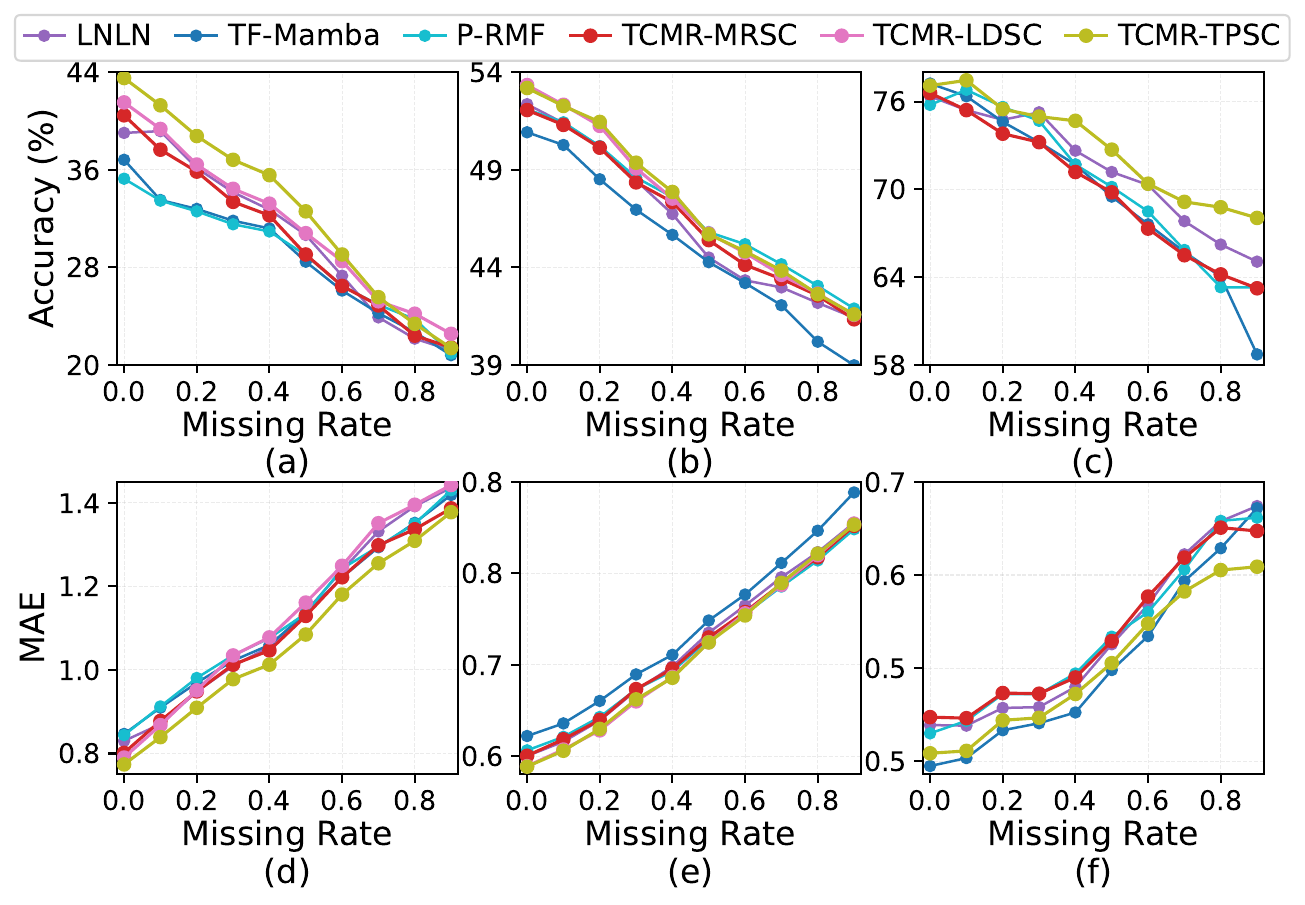}
    \caption{Performance curves under various missing rates. 
(a)--(c): Accuracy on MOSI (Acc-7), MOSEI (Has0 Acc), and SIMS (Acc-2). (d)--(f): MAE on MOSI, MOSEI, and SIMS.}
    \label{fig:missing_rate_results}
\end{figure}

\paragraph{Overall Performance}
Tables~\ref{tab:main_results}–\ref{tab:sims_results} present the evaluation results of various models on three MSA benchmark datasets. On the MOSI dataset, TCMR-TPSC achieves state-of-the-art performance across all evaluation metrics. In particular, TCMR-TPSC improves Acc-5 by 6.68\% and reduces MAE by 6.37\% compared to LNLN. Among the \texttt{TCMR-} variants, TCMR-TPSC consistently achieves the best performance across all datasets. This is because TCMR-MRSC struggles to capture the actual semantic loss from the text, as it relies solely on the missing rate. TCMR-LDSC, on the other hand, estimates completeness based on sentiment scores assigned to individual tokens, which does not consider contextual semantics and can lead to inaccurate completeness labels. These results further highlight the importance of semantic-aware completeness estimation. Further analyses with competitive baselines are provided in Appendix~\ref{sec:analyses_baseline}.

\begin{table*}[t!]
\centering
\setlength{\tabcolsep}{2.5pt}
\caption{A comprehensive ablation study of the proposed TCMR framework on the MOSI and MOSEI datasets.}
\label{tab:ablation_combined}
\resizebox{\textwidth}{!}{%
\begin{tabular}{l|cccccc|cccccc}
\toprule
\multirow{2}{*}{Method} & \multicolumn{6}{c|}{\textbf{MOSI}} & \multicolumn{6}{c}{\textbf{MOSEI}} \\
 & Acc-7 & Acc-5 & Non0 Acc / F1 & Has0 Acc / F1 & MAE & Corr & Acc-7 & Acc-5 & Non0 Acc / F1 & Has0 Acc / F1 & MAE & Corr \\
\midrule
\multicolumn{13}{l}{w/o AOS} \\
\quad CompNet-first   
& 27.67 & 30.65 & 66.88 / 66.84 & 66.45 / 66.29 & 1.1671 & 43.06 
& 43.33 & 43.48 & 65.37 / 64.09 & 69.14 / 65.16 & 0.7827 & 32.61 \\

\quad End2End     
& 29.68 & 32.07 & 68.15 / 68.04 & 67.58 / 67.36 & 1.1849 & 49.15 
& 46.82 & 47.75 & 77.97 / \textbf{77.43} & 76.27 / 76.44 & 0.6619 & \textbf{58.71} \\

\quad CompNet-later  
& 32.24 & 36.17 & 71.16 / 71.22 & 70.41 / 70.37 & 1.0944 & 50.91 
& 43.11 & 43.86 & 73.75 / 71.83 & 74.58 / 73.47 & 0.7547 & 46.85 \\

\midrule
w/o IPFG          
& 31.31 & 34.81 & 69.02 / 68.91 & 68.65 / 68.42 & 1.1674 & 45.89 
& 46.05 & 47.06 & 75.92 / 75.80 & 72.71 / 73.45 & 0.6816 & 57.22 \\

\midrule
\textbf{TCMR (Full)} 
& \textbf{32.80} & \textbf{36.73} & \textbf{72.01 / 71.67} & \textbf{70.93 / 70.49} & \textbf{1.0721} & \textbf{52.28} 
& \textbf{47.27} & \textbf{48.16} & \textbf{77.99} / 76.99 & \textbf{77.59 / 77.21} & \textbf{0.6614} & 58.63 \\

\bottomrule
\end{tabular}
}
\end{table*}

On the MOSEI and SIMS datasets, TCMR-TPSC consistently shows strong performance across most evaluation metrics. In particular, on MOSEI, it achieves the best results on Has0 Acc / F1, MAE, and Corr, while remaining competitive on Acc-7, Acc-5, and Non0 Acc / F1. On SIMS, it achieves the best results on Acc-3, Acc-2, MAE, and Corr. While TCMR-TPSC shows slightly lower performance on a few classification metrics compared to CENet and P-RMF, this can be attributed to the discretization involved in computing such metrics. For instance, Acc-7 converts continuous sentiment predictions into discrete classes via rounding. When a prediction lies close to a rounding boundary, it may fall into a different class than the ground truth, leading to misclassification despite a lower regression error. For this reason, regression metrics more accurately reflect the precision of sentiment prediction. A more detailed analysis is provided in Appendix~\ref{sec:metric_level_comp}.

To provide a comprehensive comparison with reconstruction-based approaches, we plot performance changes across missing rates in Figure~\ref{fig:missing_rate_results}. As shown in the figure, TCMR-TPSC consistently outperforms baseline methods, even as the missing rate increases. These results support our motivation that accurately estimating semantic information loss in incomplete text enables effective reconstruction of missing semantics, thereby leading to improved overall performance.
\begin{table}[t]
\centering
\small
\setlength{\tabcolsep}{3pt}
\caption{A comprehensive ablation study of the proposed TCMR framework on the SIMS dataset.}
\label{tab:ablation_sims}
\resizebox{\columnwidth}{!}{%
\begin{tabular}{l|cccccc}
\toprule
Method & Acc-5 & Acc-3 & Acc-2 & F1 & MAE & Corr \\
\midrule
\multicolumn{7}{l}{w/o AOS} \\

\quad CompNet-first
& 22.92 & 53.79 & 69.82 & 60.81 & 0.5764 & 0.171 \\

\quad End2End
& 29.26 & 53.92 & 68.57 & 67.71 & 0.5374 & 0.321 \\

\quad CompNet-later
& 29.71 & 55.49 & 70.04 & 68.48 & 0.5416 & 0.339 \\

\midrule
w/o IPFG
& 30.15 & 55.57 & 70.09 & 66.93 & 0.5451 & 0.336 \\

\midrule
\textbf{TCMR (Full)}
& \textbf{31.15} & \textbf{57.45} & \textbf{72.81} & \textbf{68.88} & \textbf{0.5212} & \textbf{0.379} \\

\bottomrule
\end{tabular}%
}

\end{table}

\section{Ablation Study}
As shown in Table~\ref{tab:ablation_combined} and Table~\ref{tab:ablation_sims}, we conduct an ablation study on the MOSI, MOSEI, and SIMS datasets to analyze the contribution of each component. In the w/o AOS setting, we design several alternative multi-task optimization strategies. In \texttt{CompNet-first}, we first optimize $\Theta^{\text{comp}}$, and then optimize the remaining modules while freezing both the shared BERT encoder and the trained \texttt{CompNet}. In contrast, \texttt{CompNet-later} first optimizes $\Theta^{\text{other}}$ using the completeness label $w$, and then optimizes $\theta^{\text{comp}}$ while freezing the shared BERT encoder. \texttt{End2End} jointly optimizes all model parameters without any staged optimization. In the w/o IPFG setting, we replace IPFG with a simpler proxy feature generator following prior work~\cite{zhang2024robust}. Specifically, instead of generating modality-specific proxy features and combining them through a gating mechanism, the audio and vision features are directly used as inputs to a single network that produces the proxy feature.

According to Table~\ref{tab:ablation_combined}, \texttt{CompNet-first} consistently achieves the lowest performance across all datasets. One possible reason is that freezing the shared BERT encoder after completeness learning prevents it from being further adapted for sentiment prediction, leading to misalignment between the two objectives. This result highlights the necessity of continuously updating the shared BERT encoder across both tasks, while preserving the hierarchical structure from completeness learning to sentiment learning. In contrast, \texttt{CompNet-later} achieves relatively competitive performance with \texttt{TCMR (Full)} on MOSI, while it performs relatively worse on MOSEI. We attribute this discrepancy to the differences in scale and characteristics between the two datasets. Compared to MOSI, MOSEI contains substantially more samples and covers a wider range of topics, indicating that a broader range of sentiment expression patterns must be learned. As a result, learning completeness estimation by optimizing a CompNet with relatively few parameters after the shared BERT encoder has already been optimized for ground-truth sentiment labels may be insufficient to capture the diverse patterns present in MOSEI. In this regard, \texttt{End2End} shows relatively more competitive performance on MOSEI than on MOSI, which we attribute to MOSEI's greater diversity reducing the influence of the dominant task, thereby naturally alleviating gradient conflict. Nevertheless, its inherent end-to-end nature still results in sub-optimal performance compared to \texttt{TCMR (Full)}. Removing \texttt{IPFG} consistently degrades performance across all datasets. This suggests that it adaptively regulates the contribution of each auxiliary modality for incomplete text reconstruction, effectively filtering redundant information while enhancing complementary semantic cues.
\section{In-depth Analysis}
\label{sec:deep_analysis}

\begin{figure}[bt]
  \centering

  \begin{subfigure}{0.9\linewidth}
    \centering
    \includegraphics[width=\linewidth]{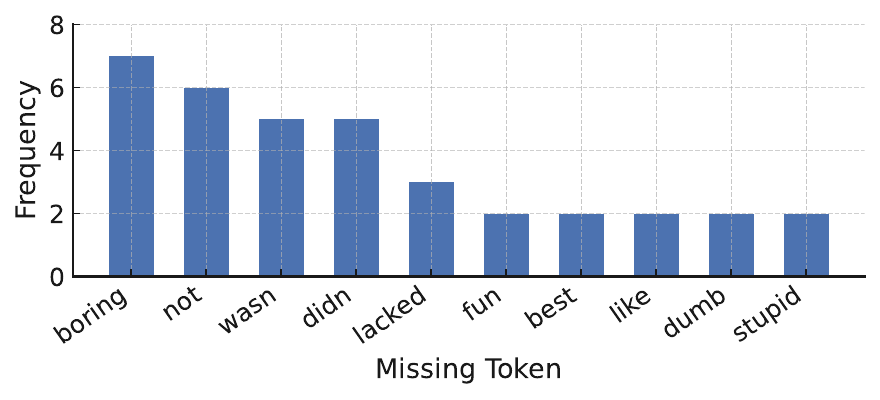}
    \caption{MOSI}
    \label{fig:mosi}
  \end{subfigure}\vfill
  \begin{subfigure}{0.9\linewidth}
    \centering
    \includegraphics[width=\linewidth]{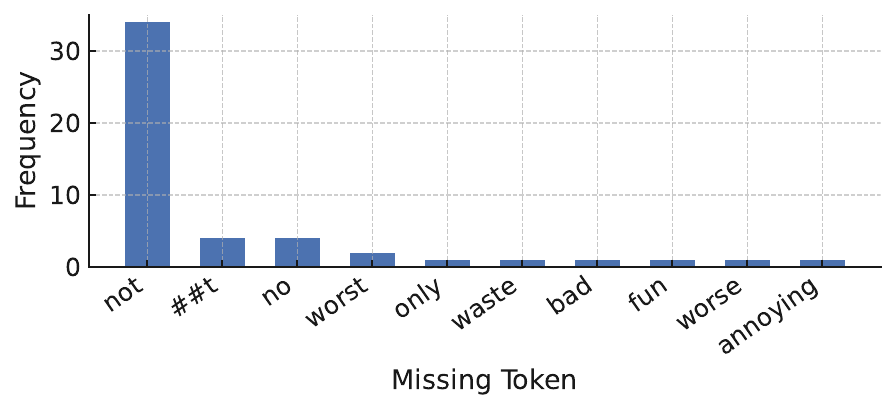}
    \caption{MOSEI}
    \label{fig:mosei}
  \end{subfigure}

  \caption{Comparison of error-triggering sentiment clue tokens across two benchmark datasets. Each histogram shows the frequency of missing tokens that cause misclassification in the sentiment classifier.}
  \label{fig:sentiment_clue}
\end{figure}

\paragraph{Sentiment Token Sensitivity}
\label{sec:sentiment_clue}
To further examine the effectiveness of the TPSC label, we conduct a sensitivity analysis on sentiment-related tokens. First, we collect samples that are correctly classified by a pretrained sentiment polarity classifier on complete textual inputs. For each sample, we construct a perturbation token set by masking each word token in turn. Then, we count the frequency of tokens whose masking causes a previously correct prediction to become incorrect (detailed procedures are provided in Appendix~\ref{sec:sensitivity_of_tpsc}).

As shown in Figure~\ref{fig:sentiment_clue}, misclassifications are concentrated on affective tokens such as [\textit{'boring'}], [\textit{'not'}], and [\textit{'fun'}]. This observation supports our hypothesis that TCP indeed reflects the preservation of sentiment-relevant information, as discussed in Section~\ref{sec:label_gen}. Consequently, \texttt{CompNet} provides a more accurate estimate of semantic completeness, enabling more reliable reconstruction by maximizing complementary semantic information from auxiliary modalities.

\begin{figure}[t]
    \centering
    \includegraphics[width=\linewidth]{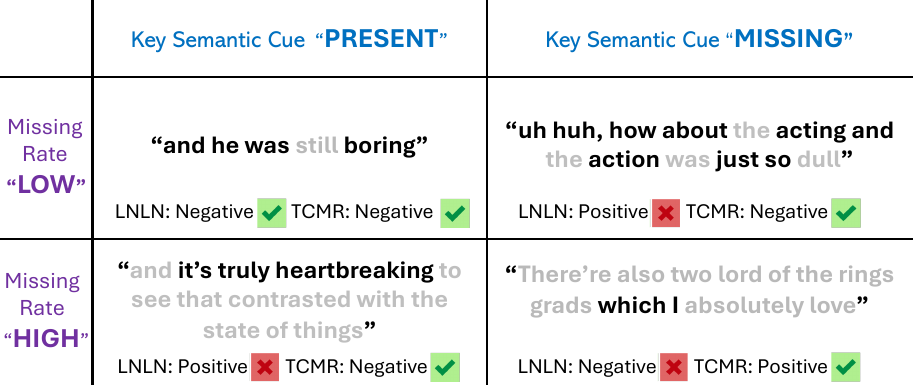}
    \caption{Qualitative comparison between the LNLN and TCMR models under varying missing-text conditions. Each cell presents an example utterance where key semantic cues are either present or missing.}
    \label{fig:case_study}
\end{figure}

\paragraph{Case Study}
Figure~\ref{fig:case_study} illustrates how LNLN and TCMR behave under different missing-text conditions. Overall, LNLN fails to accurately reconstruct missing semantics. This stems from overestimating or underestimating the semantic information remaining in incomplete text, leading to incorrect sentiment prediction. In contrast, TCMR produces correct predictions through more accurate completeness estimation. This allows TCMR to adaptively balance its reliance on auxiliary modalities based on the extent of semantic loss, resulting in a more faithful reconstruction. Notably, the bottom-right cell provides an interesting insight. Although LNLN can estimate completeness accurately in this case, due to the high missing rate and the loss of a key sentiment token, it still fails to produce the correct sentiment prediction. This suggests that inaccurate completeness estimation during training may limit LNLN's ability to reconstruct missing semantics. Therefore, accurately estimating the semantic information retained in incomplete text is crucial not only for effective reconstruction but also for stable sentiment prediction.

\begin{table*}[t]
\centering
\small
\caption{Qualitative comparison of completeness labels. 
Each case shows how MRSC, LDSC, and TPSC assign completeness labels depending on whether sentiment-related tokens are missing. In the TPSC column, the value in parentheses denotes $P(y_{sp}\mid\text{complete})$, while the value above it denotes $P(y_{sp}\mid\text{incomplete})$. 
\cmark\ and \xmark\ indicate correct and incorrect completeness estimation, respectively.}
\label{tab:qualitative_ldsc}

\renewcommand{\arraystretch}{1.35}
\setlength{\tabcolsep}{4pt}

\begin{tabular}{|c|>{\centering\arraybackslash}m{4.8cm}|>{\centering\arraybackslash}m{4.8cm}|c|c|c|}
\hline
\textbf{Case} & \textbf{Complete Text} & \textbf{Incomplete Text} & \textbf{MRSC} & \textbf{LDSC} & \textbf{TPSC} \\
\hline

\textbf{1}
& I was happy to see it
& I \texttt{[UNK]} happy to see it
& 0.8333\,\cmark
& 0.9888\,\cmark
& \begin{tabular}{c}
0.9526\\[-2pt]
{\scriptsize (0.9580)}
\end{tabular}\,\cmark \\
\hline

\textbf{2}
& The soundtrack is good its really good some really great songs
& \texttt{[UNK]} soundtrack \texttt{[UNK]} \texttt{[UNK]} \texttt{[UNK]} really \texttt{[UNK]} \texttt{[UNK]} \texttt{[UNK]} \texttt{[UNK]} \texttt{[UNK]}
& 0.1818\,\cmark
& 0.0044\,\cmark
& \begin{tabular}{c}
0.3077\\[-2pt]
{\scriptsize (0.9665)}
\end{tabular}\,\cmark \\
\hline

\textbf{3}
& But it was really really awesome
& but it was \texttt{[UNK]} \texttt{[UNK]} \texttt{[UNK]}
& 0.5000\,\xmark
& 0.0112\,\cmark
& \begin{tabular}{c}
0.3179\\[-2pt]
{\scriptsize (0.9680)}
\end{tabular}\,\cmark \\
\hline

\textbf{4}
& Kids are gonna love the film
& \texttt{[UNK]} \texttt{[UNK]} \texttt{[UNK]} love the \texttt{[UNK]}
& 0.3333\,\xmark
& 0.9808\,\cmark
& \begin{tabular}{c}
0.8638\\[-2pt]
{\scriptsize (0.9571)}
\end{tabular}\,\cmark \\
\hline

\textbf{5}
& I will admit I'm a big Johnny Depp fan
& I will \texttt{[UNK]} I'm \texttt{[UNK]} big Johnny Depp fan
& 0.7931\,\cmark
& 0.0690\,\xmark
& \begin{tabular}{c}
0.9160\\[-2pt]
{\scriptsize (0.9491)}
\end{tabular}\,\cmark \\
\hline

\textbf{6}
& and you're better off saving your money uh and maybe renting it when it comes out on uh video so
& and \texttt{[UNK]} better \texttt{[UNK]} saving \texttt{[UNK]} money \texttt{[UNK]} \texttt{[UNK]} maybe renting it when it \texttt{[UNK]} out \texttt{[UNK]} uh \texttt{[UNK]} \texttt{[UNK]}
& 0.5921\,\cmark
& 0.9928\,\xmark
& \begin{tabular}{c}
0.4608\\[-2pt]
{\scriptsize (0.6153)}
\end{tabular}\,\cmark \\
\hline

\textbf{7}
& I really dig this movie
& I \texttt{[UNK]} \texttt{[UNK]} this \texttt{[UNK]}
& 0.4000\,\xmark
& 0.5000\,\xmark
& \begin{tabular}{c}
0.1006\\[-2pt]
{\scriptsize (0.9477)}
\end{tabular}\,\cmark \\
\hline

\end{tabular}
\end{table*}

\paragraph{Comparison of Completeness Labels}
Table~\ref{tab:qualitative_ldsc} provides a qualitative comparison of different completeness labeling strategies under random missing conditions. Cases 3 and 4 highlight the limitations of MRSC. Since MRSC estimates completeness solely based on the proportion of missing tokens, it cannot capture the actual semantic information loss in the incomplete text. In contrast, LDSC and TPSC more appropriately reflect completeness by considering the semantic importance of sentiment-related tokens. Cases 5–7 reveal the limitations of LDSC. For example, in Cases 5 and 7, words such as [\textit{'fan'}] and [\textit{'dig'}] are used as positive sentiment expressions. However, as SentiWordNet assigns sentiment scores based on the most frequent sense of a word, the assigned score may differ from the sense used in the actual context. Moreover, in Case 6, [\textit{'better'}] is assigned a positive sentiment score in SentiWordNet, leading LDSC to yield a high completeness value even though the overall sentence conveys negative sentiment. In contrast, TPSC produces relatively correct completeness in these challenging cases due to its ability to capture sentiment based on the context of the full sentence rather than individual words.

\section{Conclusion}
\label{sec:conclusion}
In this work, we reinterpret semantic information loss in incomplete data based on the change in target-class probability between complete and incomplete data to address the performance degradation of text-centric fusion models in MSA caused by noisy or missing data. First, we generate pseudo-labels that provide supervision for training a completeness estimator. We further propose an optimization strategy that mitigates gradient conflicts between the completeness estimation and sentiment prediction tasks, thereby stabilizing hierarchical multi-task learning. We believe the proposed completeness estimation approach can be broadly applied beyond MSA, such as confidence-aware decision making and low-quality data detection or restoration.

\section*{Limitations}
Although TCMR consistently outperforms existing
reconstruction-based MSA models, several limitations remain to be addressed in future work. First, this study only focuses on reconstructing semantic information loss in the textual modality. Extending the TPSC-based completeness estimation to other modalities, such as audio and vision, could improve general multimodal fusion approaches beyond text-centric fusion. Second, TCMR relies on pseudo-labeling, which inherently introduces potential noise in the supervision signal. Future work could mitigate this issue by adopting confidence-aware filtering, using only high-confidence TPSC predictions as pseudo-labels while relying on complementary strategies such as LDSC for less confident samples. For samples where both TPSC and LDSC yield low confidence, human verification could be incorporated to further reduce label noise.

\section*{Acknowledgments}
This work was supported by Institute of Information \& Communications Technology Planning \& Evaluation (IITP) grant funded by the Korea government (MSIT) (No. RS-2022-II220608, RS-2022-II220320, and RS-2024-00398115); and the Korea Electronics Technology Institute (KETI) through the project ``Development of Modular Humanoid Physical AI Technology for Performing Complex Tasks in Industrial Environments.''





\bibliography{custom}


\appendix

\clearpage

\section{Implementation Details}
\label{sec:imple_details}

\subsection{Dataset Statistics}
\label{sec:data_stas}
\vspace{-1.0em}
\begin{table}[H]
\centering
\small
\caption{Statistics of the MOSI, MOSEI, and SIMS datasets used in our experiments.}
\label{tab:dataset_stats}
\setlength{\tabcolsep}{4pt}
\renewcommand{\arraystretch}{1.15}
\resizebox{\columnwidth}{!}{%
\begin{tabular}{lccc}
\toprule
 & \textbf{MOSI} & \textbf{MOSEI} & \textbf{SIMS} \\
\midrule
\#Samples & 2,199 & 22,856 & 2,281 \\
Train / Val / Test & 1,284 / 229 / 686 & 16,326 / 1,871 / 4,659 & 1,368 / 456 / 457 \\
\#Speakers & 93 & 1,000+ & -- \\
\#Topics & 89 & 250+ & -- \\
Source & YouTube & YouTube & Movies / TV \\
Annotation Type & Utterance-level & Utterance-level & Utterance-level \\
Sentiment Score & --3 to +3 & --3 to +3 & --1 to +1 \\
\bottomrule
\end{tabular}%
}
\end{table}

Table~\ref{tab:dataset_stats} provides key statistics of the three benchmark datasets used in our experiments. Among them, MOSEI stands out as substantially larger and more diverse in speakers and topics than MOSI and SIMS, resulting in greater variability across acoustic, visual, and textual modalities. Due to this diversity, MOSEI is considered a more challenging benchmark than MOSI and SIMS.

\subsection{Model Configuration}
The proposed TCMR model was implemented in PyTorch (v2.2.1) with Python 3.11.7 and trained on a workstation with an NVIDIA GeForce RTX 4090 GPU and an AMD Ryzen Threadripper PRO 5975WX CPU. To offer a thorough specification of our architecture design, Table~\ref{tab:model_parameters} provides the detailed configurations of all modules within TCMR.

\begin{table*}[t!]
    \centering
    \small
    \setlength{\tabcolsep}{5pt}
    \caption{Network configurations of TCMR. Transformer-based modules list the number of layers, sequence lengths, token counts, input dimensions, attention heads, and hidden dimensions for both MOSI and MOSEI (shown as MOSI / MOSEI). MLP-based modules display the input dimensions and hidden-layer widths used in each component.}
    \label{tab:arch_all}
    \resizebox{\textwidth}{!}{%
    \begin{tabular}{lccccccccc}
        \toprule
        \textbf{Module (MOSI / MOSEI)} 
        & \textbf{Type} 
        & \textbf{Notation} 
        & \textbf{\#Layers} 
        & \textbf{Seq. len} 
        & \textbf{Token len} 
        & \textbf{Input dim} 
        & \textbf{\#Heads} 
        & \textbf{Hidden dim} \\
        \midrule\midrule
        \multicolumn{9}{l}{\textbf{Transformer-based modules}} \\
        \midrule\midrule
        BERT Encoder 
            & Transformer 
            & $\theta^{\text{bert}}$ 
            & 12 
            & 50 / 50
            & 50 
            & 768 / 768
            & 12 
            & 768 \\
        Text Encoder 
            & Transformer 
            & $\theta^{\text{trans}}_{t}$ 
            & 2 
            & 49 / 49
            & 8 
            & 768 / 768
            & 8 
            & 128 \\
        Audio Encoder 
            & Transformer 
            & $\theta^{\text{trans}}_{a}$ 
            & 2 
            & 375 / 500
            & 8 
            & 5 / 74
            & 8 
            & 128 \\
        Vision Encoder
            & Transformer 
            & $\theta^{\text{trans}}_{v}$ 
            & 2 
            & 500 / 500
            & 8 
            & 20 / 35
            & 8 
            & 128 \\
        A2T Generator (audio $\rightarrow$ text) 
            & Transformer 
            & $\theta^{\text{gen}}_{a}$ 
            & 2 
            & 16 / 16
            & -- 
            & 128 / 128
            & 8 
            & 128 \\
        V2T Generator (vision $\rightarrow$ text) 
            & Transformer 
            & $\theta^{\text{gen}}_{v}$ 
            & 2 
            & 16 / 16
            & -- 
            & 128 / 128
            & 8 
            & 128 \\
        Text Refinement Transformer 
            & Transformer 
            & $\theta^{\text{ref}}_{t}$ 
            & 2 
            & 8 / 8
            & -- 
            & 128 / 128
            & 8 
            & 128 \\
        CrossTransformer (fusion head) 
            & Transformer 
            & $\theta^{\text{cross}}$ 
            & 2 
            & 8 / 8
            & -- 
            & 128 / 128
            & 8 
            & 128 \\
        Reconstructor (Audio)
            & Transformer 
            & $\theta^{\text{recon}}_{a}$ 
            & 2 
            & 8 / 8
            & -- 
            & 128 / 128
            & 8 
            & 128 \\
        Reconstructor (Vision) 
            & Transformer 
            & $\theta^{\text{recon}}_{v}$ 
            & 2 
            & 8 / 8
            & -- 
            & 128 / 128
            & 8 
            & 128 \\
        \midrule\midrule
        \multicolumn{9}{l}{\textbf{MLP-based modules}} \\
        \midrule\midrule
        Completeness Estimator (CompNet) 
            & MLP 
            & $\theta^{\text{comp}}$ 
            & 6 
            & -- 
            & -- 
            & 768 / 768
            & -- 
            & [768, 768, 1536, 768, 384, 1] \\
        Polarity Classifier (Classifier)
            & MLP 
            & $\theta^{\text{classifier}}_{\text{pre}}$ 
            & 2 
            & -- 
            & -- 
            & 768 / 768
            & -- 
            & [384, 3] \\
        Gating Network (GateNet) 
            & MLP 
            & $\theta^{\text{gate}}$ 
            & 2 
            & -- 
            & -- 
            & 256 / 256
            & -- 
            & [128, 1] \\
        Text Regressor (Regressor)
            & MLP 
            & $\theta^{\text{regressor}}$ 
            & 1 
            & -- 
            & -- 
            & 128 / 128
            & -- 
            & [1] \\
        \bottomrule
    \end{tabular}
    }%
\label{tab:model_parameters}
\end{table*}

\label{sec:network_parameters}

\begin{table}[t!]
    \centering
    \caption{Performance sensitivity to sentiment class granularity ($k$) in TPSC label generation.}
    \label{tab:TPSC_cls_results}

    \begin{subtable}{\columnwidth}
        \centering
        \caption{MOSI dataset}
        \label{tab:mosi_sub}
        \resizebox{\linewidth}{!}{%
        \begin{tabular}{lcccccc}
            \toprule
            \textbf{Model} & \textbf{Acc-7} & \textbf{Acc-5} & \textbf{Non0 Acc/F1} & \textbf{Has0 Acc/F1} & \textbf{MAE} & \textbf{Corr} \\
            \midrule
            TPSC-2cls & 31.45 & 36.01 & 70.01 / 69.91 & 69.46 / 69.25 & 1.1073 & 50.66 \\
            TPSC-3cls & \textbf{32.80} & \textbf{36.73} & \textbf{72.01 / 71.67} & \textbf{70.93 / 70.49} & \textbf{1.0721} & \textbf{52.28} \\
            TPSC-5cls & 30.96 & 35.12 & 69.51 / 69.09 & 68.93 / 68.42 & 1.1449 & 49.62 \\
            TPSC-7cls & 32.16 & 35.35 & 70.85 / 70.76 & 69.74 / 69.54 & 1.1040 & 49.28 \\
            \bottomrule
        \end{tabular}%
        }
    \end{subtable}

    \vspace{0.4cm}

    \begin{subtable}{\columnwidth}
        \centering
        \caption{MOSEI dataset}
        \label{tab:mosei_sub}
        \resizebox{\linewidth}{!}{%
        \begin{tabular}{lcccccc}
            \toprule
            \textbf{Model} & \textbf{Acc-7} & \textbf{Acc-5} & \textbf{Non0 Acc/F1} & \textbf{Has0 Acc/F1} & \textbf{MAE} & \textbf{Corr} \\
            \midrule
            TPSC-2cls & 46.41 & 47.25 & 77.76 / 77.32 & 76.31 / 76.53 & 0.6677 & 57.95 \\
            TPSC-3cls & \textbf{47.27} & \textbf{48.16} & \textbf{77.99} / 76.99 & \textbf{77.59 / 77.21} & \textbf{0.6614} & 58.63 \\
            TPSC-5cls & 46.94 & 47.96 & 77.79 / \textbf{77.56} & 75.46 / 75.95 & 0.6638 & \textbf{59.28} \\
            TPSC-7cls & 46.75 & 47.63 & 77.74 / 76.74 & 77.15 / 76.79 & 0.6667 & 57.81 \\
            \bottomrule
        \end{tabular}%
        }
    \end{subtable}

    \vspace{0.4cm}

    \begin{subtable}{\columnwidth}
        \centering
        \caption{SIMS dataset}
        \label{tab:sims_sub}
        \resizebox{\linewidth}{!}{%
        \begin{tabular}{lcccccc}
            \toprule
            \textbf{Model} & \textbf{Acc-5} & \textbf{Acc-3} & \textbf{Acc-2} & \textbf{F1} & \textbf{MAE} & \textbf{Corr} \\
            \midrule
            TPSC-2cls & 31.15 & \textbf{57.45} & \textbf{72.81} & 68.88 & \textbf{0.5212} & \textbf{0.379} \\
            TPSC-3cls & 32.43 & 56.23 & 71.72 & \textbf{68.96} & 0.5478 & 0.321 \\
            TPSC-5cls & \textbf{33.14} & 55.45 & 69.13 & 68.06 & 0.5465 & 0.329 \\
            \bottomrule
        \end{tabular}%
        }
    \end{subtable}
\end{table}

\subsection{Evaluation Metrics}
\label{sec:eval_metrics}
Following prior work~\cite{zhang2024robust}, we adopt the same evaluation metrics.
For MOSI and MOSEI, we report both classification metrics (Acc-7, Acc-5, Non0 Acc / F1, Has0 Acc / F1) and regression metrics (MAE, Corr). The classification metrics are computed by discretizing the continuous sentiment scores into predefined sentiment intervals. For SIMS, we report Acc-2, Acc-3, Acc-5, F1, MAE, and Corr. Non0 Acc denotes negative/positive classification (excluding zero-valued samples), whereas Has0 Acc denotes negative/non-negative classification, where zero-valued samples are treated as non-negative.

\subsection{Multimodal Fusion Module}
\label{sec:fusion}

The multimodal fusion module follows the design of prior work~\cite{zhang2024robust}, and its key components are briefly summarized here. Starting from the reconstructed text representation $\hat H_t$, a refinement Transformer encoder applies self-attention, producing layerwise refined features $H_t^{+(i)}$ with $i \in \{1, \ldots, l^{\text{ref}}\}$. Based on $H_t^{+(i)}$, cross-modal attention integrates complementary cues from vision and audio across multiple layers. At each layer, the current refined text feature $H_t^{+(i)}$ serves as the query and attends to $H_v$ and $H_a$, while the fused representation updates $H_f^{(i)}$ through residual accumulation. In compact form, $H_f^{(i)} = H_f^{(i-1)} + \mathrm{MHA}(H_t^{+(i)}, H_v) + \mathrm{MHA}(H_t^{+(i)}, H_a)$ where $H_f^{0}$ is a learnable embedding and $\mathrm{MHA}(Q,K)$ denotes multi-head attention with query \(Q\) and key and value from $K$. A CrossTransformer then models interactions between $[\,H_f^{(l^{\text{ref}})},\, H_t^{+(l^{\text{ref}})}\,]$, and a regression head outputs the final sentiment prediction $\hat y$.

\section{Sentiment Polarity Classes}
\subsection{Definition of Sentiment 
Polarity Classes}
\label{sec:definition_of_sp}
To utilize TCP, we convert continuous sentiment scores into discrete polarity classes. As the score ranges differ across datasets, the class boundaries are defined separately for SIMS and MOSI/MOSEI.

For the SIMS dataset, polarity classes are defined using fixed interval thresholds. Specifically, the 2-class setting divides the scores into $[-1.0, 0.0]$ and $(0.0, 1.0]$. The 3-class setting introduces a neutral region, resulting in $[-1.0, -0.1]$, $(-0.1, 0.1]$, and $(0.1, 1.0]$. The 5-class setting adopts a finer partition given by $[-1.0, -0.7]$, $(-0.7, -0.1]$, $(-0.1, 0.1]$, $(0.1, 0.7]$, and $(0.7, 1.0]$.

For the MOSI and MOSEI datasets, polarity classes are defined by mapping sentiment scores into discrete levels. The 2-class setting separates negative ($<0$) and non-negative ($\geq 0$) samples. The 3-class, 5-class, and 7-class settings correspond to $\{-1, 0, 1\}$, $\{-2, -1, 0, 1, 2\}$, and $\{-3, -2, -1, 0, 1, 2, 3\}$, respectively.

\subsection{Optimal Choice of Class Granularity}
\label{TPSC_cls}

We conduct comparative experiments with $k \in \{2, 3, 5, 7\}$ (MOSI/MOSEI) and $k \in \{2, 3, 5\}$ (SIMS) and select k = 3 (MOSI/MOSEI) and k = 2 (SIMS) based on validation performance. As shown in Table~\ref{tab:TPSC_cls_results}, this configuration also yields the best test-set performance.

\section{Gradient Conflict Analysis}
\label{sec:aos_appendix}

In multi-task learning, jointly optimizing multiple losses often leads to gradient conflicts, resulting in biased optimization where certain tasks dominate while others are under-optimized. To address this problem, prior work~\cite{chen2018gradnorm, yu2020gradient} has proposed methods that mitigate gradient conflicts by adjusting gradient magnitudes or directions during training. For example, \citet{chen2018gradnorm} propose GradNorm, a method that automatically balances multiple tasks during training by adjusting the magnitudes of gradients derived from each task-specific loss, ensuring that all tasks learn at a similar rate. \citet{yu2020gradient} propose PCGrad, which reduces gradient interference by projecting conflicting gradients onto the normal plane of other task gradients.

\begin{table}[t]
\centering
\caption{Gradient conflict rate (GCR) analysis measured during training. F50/L50 denote the first and last 50\% of training epochs, respectively.}
\label{tab:gradient_conflict}

\setlength{\tabcolsep}{4pt}
\renewcommand{\arraystretch}{1.05}

\begin{tabular}{lcccc}
\toprule
\multirow{2}{*}{Method} 
& \multicolumn{3}{c}{GCR (\%) $\downarrow$} 
& \multirow{2}{*}{$\mathcal{L}_{comp}$ $\downarrow$} \\
\cmidrule(lr){2-4}
& All & F50 & L50 &  \\
\midrule
E2E      & 51.27 & 51.16 & 54.36 & 0.1634 \\
GradNorm & 54.84 & 56.55 & 53.16 & 0.1294 \\
PCGrad   & 52.33 & 57.43 & 48.70 & 0.0340 \\
AOS      & 48.47 & 53.67 & 43.33 & 0.0055 \\
\bottomrule
\end{tabular}

\end{table}

To validate the effectiveness of AOS, we conduct comparative experiments using existing methods as baselines. Table~\ref{tab:gradient_conflict} presents the gradient conflict rates during training and the corresponding convergence behavior of the completeness estimation loss, following the protocol of prior work~\cite{Zhang_2024_PGCM}. In joint training, $\mathcal{L}_{\text{comp}}$ tends to converge to a local optimum, which leads to an increase in GCR (see E2E in Table~\ref{tab:gradient_conflict}). While GradNorm and PCGrad mitigate gradient conflict, they are still insufficient to escape local optima due to the fundamental limitation of joint training. In contrast, AOS removes interference from other losses during the \texttt{CompNet} training phase, thereby effectively mitigating suboptimal convergence of $\mathcal{L}_{\text{comp}}$.

\section{Analysis of Completeness Labels}
\label{sec:analysis_of_tpsc}

\subsection{Lexicon-derived Completeness Label}
\label{sec:ldsc}

To further validate the effectiveness of the proposed TPSC, we introduce \textit{Lexicon-Derived Semantic Completeness} (LDSC) as a comparison baseline. LDSC measures completeness by explicitly quantifying how much sentiment-related lexical information remains in the incomplete text. Specifically, we adopt SentiWordNet~\cite{baccianella2010sentiwordnet}, a widely used lexical resource for sentiment analysis. SentiWordNet provides positive and negative sentiment scores for each word.

\begin{figure}[t]
\centering
\includegraphics[width=\columnwidth]{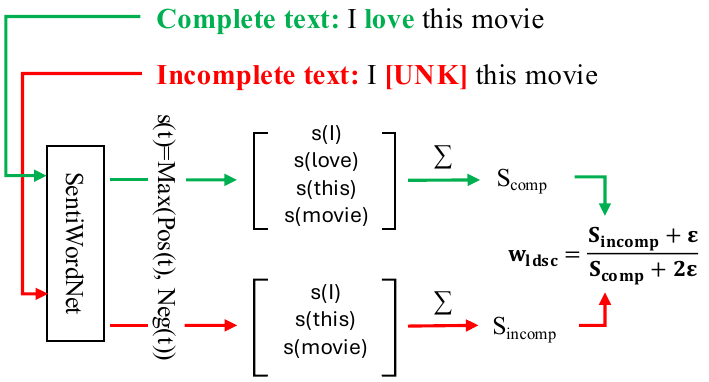}
\caption{Illustration of LDSC computation}
\label{fig:ldsc_process}
\end{figure}

Figure~\ref{fig:ldsc_process} illustrates the overall process of generating the LDSC label. First, we define the sentiment intensity score of a token $t$ as:
\begin{equation}
s(t) = \max \big( \mathrm{pos}(t), \mathrm{neg}(t) \big),
\end{equation}
where $\mathrm{pos}(t)$ and $\mathrm{neg}(t)$ denote the positive and negative sentiment scores of the token. Using this token-level score, we compute the total sentiment-related content of the complete and incomplete text:
\begin{equation}
S_{\text{comp}} = \sum_{t \in T_{\text{comp}}} s(t),
\end{equation}
\begin{equation}
S_{\text{incomp}} = \sum_{t \in T_{\text{incomp}}} s(t),
\end{equation}
where $T_{\text{comp}}$ and $T_{\text{incomp}}$ denote the token sets of the complete and incomplete text, respectively. Finally, LDSC-based completeness is computed as:
\begin{equation}
w_{ldsc} = \frac{S_{\text{incomp}} + \varepsilon}{S_{\text{comp}} + 2\varepsilon},
\end{equation}
where $\varepsilon$ is a small smoothing constant introduced to avoid numerical instability and degenerate cases in which both $S_{\text{comp}}$ and $S_{\text{incomp}}$ become zero. Intuitively, when $w_{ldsc} \approx 0$, most key sentiment tokens are missing, whereas $w_{ldsc} \approx 1$ indicates that the sentiment-related information is largely preserved.

\begin{figure}[t]
\centering

\begin{subfigure}[t]{0.32\linewidth}
    \centering
    \includegraphics[width=\linewidth]{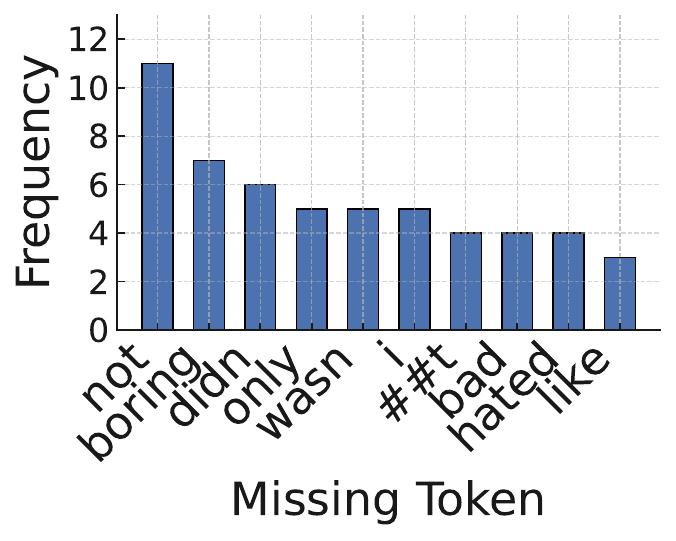}
    \caption{$c \geq 0.7$}
    \label{fig:sub1}
\end{subfigure}
\hfill
\begin{subfigure}[t]{0.32\linewidth}
    \centering
    \includegraphics[width=\linewidth]{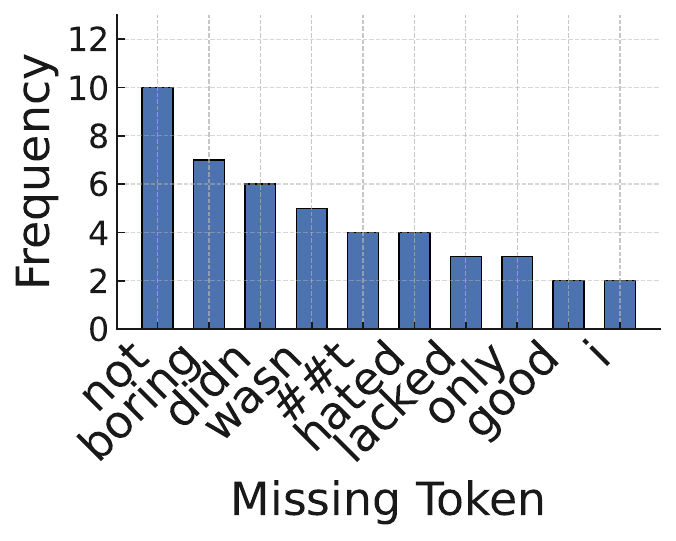}
    \caption{$c \geq 0.8$}
    \label{fig:sub2}
\end{subfigure}
\hfill
\begin{subfigure}[t]{0.32\linewidth}
    \centering
    \includegraphics[width=\linewidth]{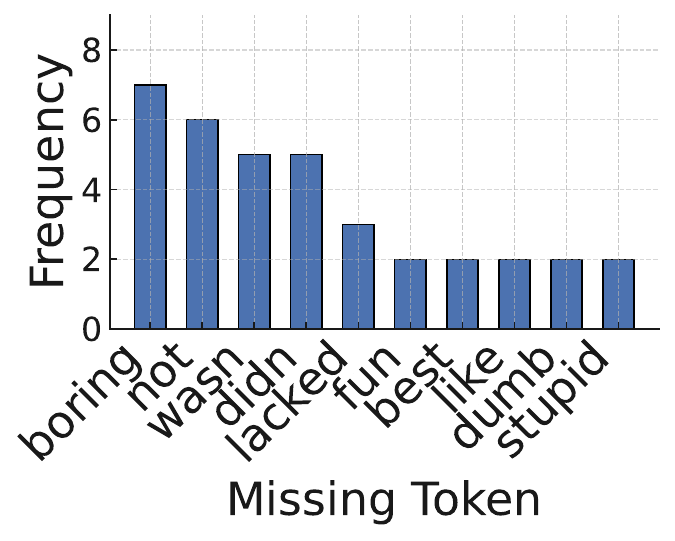}
    \caption{$c \geq 0.9$}
    \label{fig:sub3}
\end{subfigure}

\vspace{0.5em}

\begin{subfigure}[t]{0.32\linewidth}
    \centering
    \includegraphics[width=\linewidth]{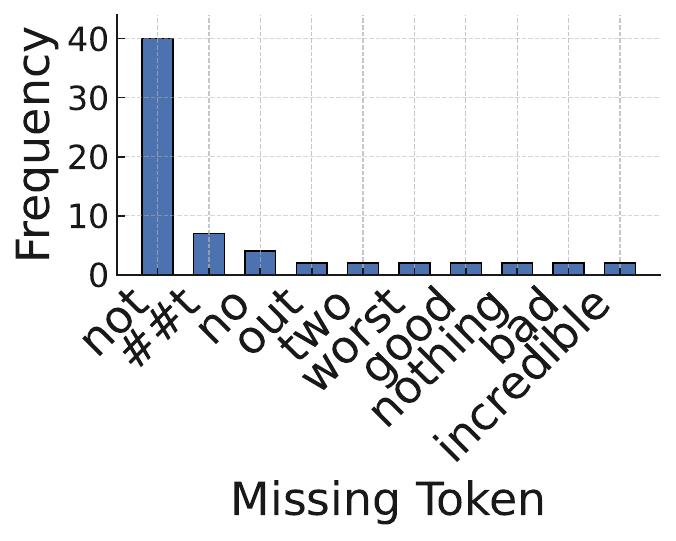}
    \caption{$c \geq 0.7$}
    \label{fig:sub4}
\end{subfigure}
\hfill
\begin{subfigure}[t]{0.32\linewidth}
    \centering
    \includegraphics[width=\linewidth]{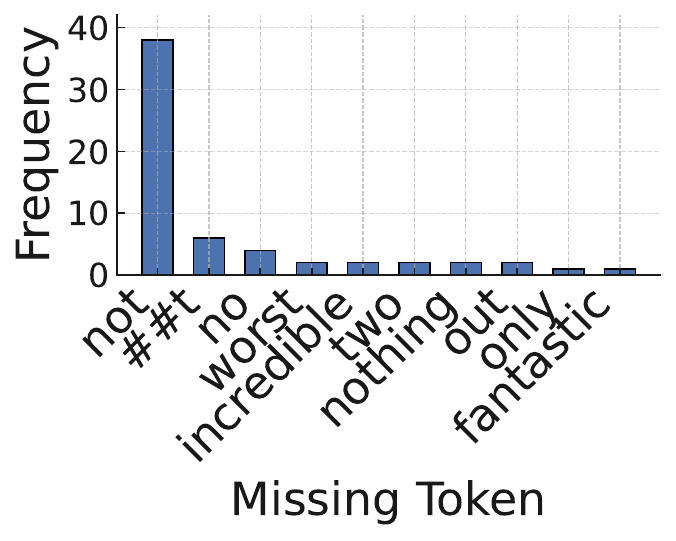}
    \caption{$c \geq 0.8$}
    \label{fig:sub5}
\end{subfigure}
\hfill
\begin{subfigure}[t]{0.32\linewidth}
    \centering
    \includegraphics[width=\linewidth]{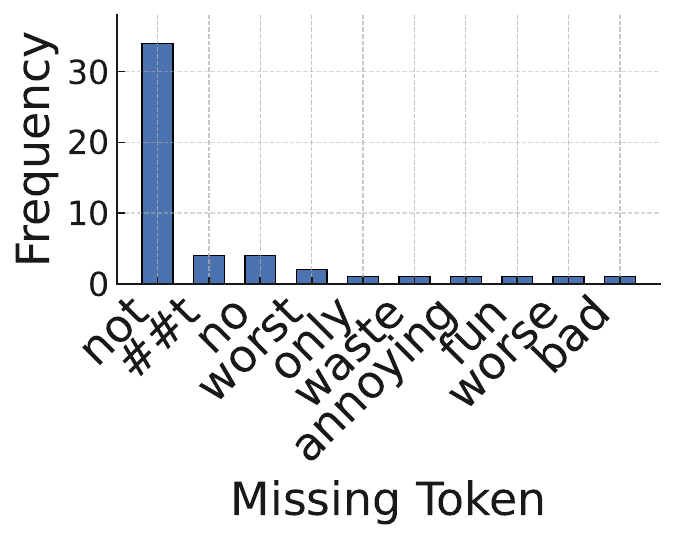}
    \caption{$c \geq 0.9$}
    \label{fig:sub6}
\end{subfigure}

\caption{Token-level sensitivity analysis across different confidence thresholds. Subfigures (a)–(c) present results on the MOSI dataset, while (d)–(f) show the corresponding results on the MOSEI dataset.}
\label{fig:token_sensitivity_threshold}
\end{figure}

\subsection{Sentiment Token Sensitivity of TPSC}
\label{sec:sensitivity_of_tpsc}
This subsection provides a more detailed description of the sentiment token sensitivity analysis introduced in Section~\ref{sec:sentiment_clue}. 

\paragraph{Analysis Procedure.}
To evaluate the sensitivity of pseudo completeness labels to key sentiment tokens, we conduct an analysis following a four-step procedure:
\begin{itemize}
    \item \textbf{Step~1: Correct-sample filtering.}  
    A pre-trained sentiment polarity classifier is applied to the complete textual inputs, and only the correctly classified samples are selected for subsequent steps. Note that a sample is included only when the true‐class probability $c \in [0,1]$ exceeds a predefined threshold.
    \item \textbf{Step~2: Token-wise Augmentation via Masking.}  
    For each correctly classified sample, we generate augmented samples by sliding a window over the sequence and removing exactly one token at each position, replacing it with \texttt{[UNK]}. For example, the sentence \emph{``I love this movie''} produces four augmented samples:
    \begin{description}[leftmargin=2em]
    \item[(1)] \texttt{[UNK]} love this movie.
    \item[(2)] I \texttt{[UNK]} this movie.
    \item[(3)] I love \texttt{[UNK]} movie.
    \item[(4)] I love this \texttt{[UNK]}.
    \end{description}

    \item \textbf{Step~3: Re-evaluation of augmented samples.}  
   Then, each augmented sample is passed through the same classifier used in Step~1, and we collect the samples in which masking a specific token causes the predicted sentiment to flip to the opposite polarity (e.g., positive $\rightarrow$ negative or negative $\rightarrow$ positive).

    \item \textbf{Step~4: Identification of error-triggering tokens.}  
    Using the polarity-flip samples collected in Step~3, we examine which token in each flipped sample triggers misclassification and compute how frequently each token leads to a misprediction.
\end{itemize}

\paragraph{Analysis Results}
Figure~\ref{fig:token_sensitivity_threshold} illustrates which tokens trigger polarity flips when masked under different thresholds. In both datasets, we consistently observed that as the threshold increases key sentiment tokens remain at the top while semantically ambiguous tokens naturally disappear. In MOSI, [\textit{‘I’}] is a sentiment-neutral token. One possible explanation is that masking it creates an ambiguous context before the verb, which may resemble negation patterns (e.g., “don’t”) learned during training and occasionally lead to prediction flips. In MOSEI, the token [\textit{'two'}] is also observed among the flip-triggering words. This is because in the original sentence \emph{``I give the movie two out of five stars''} the token [\textit{'two'}] serves as an explicit negative cue. Overall, the consistent emergence of key sentiment tokens at higher confidence thresholds suggests that the text classifier is well calibrated with respect to semantic cues.

\subsection{Upper-Bound Analysis}
\begin{table}[H]
\centering
\small
\setlength{\tabcolsep}{5pt}
\caption{Comparison between TCMR and its upper-bound variant TCMR-ub.}
\label{tab:tcmr_ub_reduced}
\begin{tabular*}{\linewidth}{@{\extracolsep{\fill}}l c c c c}
\toprule
\textbf{Method} & \textbf{Acc-5} & \textbf{Acc-2} & \textbf{MAE} & \textbf{Corr} \\
\midrule
\rowcolor{gray!15} \multicolumn{5}{c}{\textbf{\textit{MOSI Dataset}}} \\
\midrule
TCMR & 36.73 & 70.93 & 1.0721 & 52.28 \\
TCMR-ub & 41.24 & 77.57 & 0.9305 & 65.03 \\
\midrule
\rowcolor{gray!15} \multicolumn{5}{c}{\textbf{\textit{MOSEI Dataset}}} \\
\midrule
TCMR & 48.16 & 77.59 & 0.6614 & 58.63 \\
TCMR-ub & 58.72 & 79.83 & 0.5697 & 71.78 \\
\midrule
\rowcolor{gray!15} \multicolumn{5}{c}{\textbf{\textit{SIMS Dataset}}} \\
\midrule
TCMR & 31.15 & 72.81 & 0.5212 & 37.91 \\
TCMR-ub & 40.39 & 73.69 & 0.3991 & 66.47 \\
\bottomrule
\end{tabular*}
\end{table}

To validate the effectiveness of the pseudo-labels, we compare the performance of TCMR-ub, trained with target TPSC labels. Specifically, TCMR-ub is trained by directly using the pseudo-labels instead of training the confidence estimator, enabling us to assess the upper-bound performance of our framework. As shown in Table~\ref{tab:tcmr_ub_reduced}, TCMR-ub consistently outperforms TCMR across all metrics. This indicates that the pseudo-labels provide a meaningful and effective signal for guiding the reconstruction process.

\subsection{Failure Cases of TPSC}
\begin{table}[H]
\centering
\small
\begin{tabular}{lcc}
\toprule
\textbf{Dataset} & \textbf{Accuracy (\%)} & \textbf{Error Rate (\%)} \\
\midrule
MOSI  & 74.92 & 25.08 \\
MOSEI & 66.98 & 33.02 \\
SIMS  & 79.07 & 20.93 \\
\bottomrule
\end{tabular}
\caption{Accuracy of the pretrained sentiment polar- ity classifier used for TPSC pseudo-labeling on each dataset’s complete text. The higher error rate indicates a greater likelihood of noisy pseudo-labels.}
\label{tab:tpsc_classifier_accuracy}
\end{table}

\begin{table*}[t]
\centering
\small
\setlength{\tabcolsep}{5pt}
\resizebox{\textwidth}{!}{%
\begin{tabular}{llccc}
\toprule
\textbf{Pattern} & \textbf{Utterance} & \textbf{GT} & \textbf{Pred} & \textbf{TCP} \\
\midrule
\multirow{2}{*}{Ambiguous}
 & There are some funny moments & Neutral & Positive & \textbf{0.038} \\
 & Um I did enjoy it & Neutral & Positive & \textbf{0.027} \\
\midrule
\multirow{2}{*}{Sarcasm}
 & Hi I'm pretty I have a giant smile I'm supposed to know things um walk of screen & Negative & Positive & \textbf{0.131} \\
 & He um had all the charm of a narcissist xxx boy the whole film & Negative & Positive & \textbf{0.022} \\
\bottomrule
\end{tabular}%
}
\caption{Representative failure cases of the pretrained classifier in TPSC pseudo-labeling.}
\label{tab:tpsc_failure_cases}
\end{table*}

\begin{table*}[t]
\centering
\small
\setlength{\tabcolsep}{5pt}

\begin{tabular}{llccccccc}
\toprule
\textbf{Dataset} & \textbf{TCMR vs} & \textbf{Acc-7} & \textbf{Acc-5} & \textbf{Non0 A/F} & \textbf{Has0 A/F} & \textbf{MAE} & \textbf{Corr} & \textbf{W-L} \\
\midrule
\multirow{2}{*}{MOSI}
 & CENet & +3.31 & +3.83 & +2.13 / +1.73 & +1.50 / +1.10 & +0.108 & +4.04 & \textbf{8-0} \\
 & P-RMF & +3.89 & +5.49 & +2.10 / +2.10 & +1.65 / +1.65 & +0.058 & +2.21 & \textbf{8-0} \\
\midrule
\multirow{2}{*}{MOSEI}
 & CENet & $-$0.09 & $-$0.08 & +0.98 / $-$0.27 & +1.63 / +0.98 & +0.001 & +0.39 & \textbf{5-3} \\
 & P-RMF & +0.26 & +0.34 & $-$0.08 / $-$0.77 & +0.41 / +0.30 & +0.005 & +0.18 & \textbf{6-2} \\
\bottomrule
\end{tabular}

\vspace{6pt}

\begin{tabular}{llccccccc}
\toprule
\textbf{Dataset} & \textbf{TCMR vs} & \textbf{Acc-5} & \textbf{Acc-3} & \textbf{Acc-2} & \textbf{F1} & \textbf{MAE} & \textbf{Corr} & \textbf{W-L} \\
\midrule
\multirow{2}{*}{SIMS}
 & CENet & +9.54 & +3.97 & +3.58 & +11.91 & +0.109 & +0.365 & \textbf{6-0} \\
 & P-RMF & $-$1.57 & +0.15 & +2.23 & $-$1.48 & +0.022 & +0.018 & \textbf{4-2} \\
\bottomrule
\end{tabular}

\caption{Detailed comparison between TCMR-TPSC and CENet/P-RMF on MOSI, MOSEI, and SIMS.}
\label{tab:tcmr_vs_cenet_prmf}
\end{table*}

Since TPSC is based on pseudo-labeling, incorrect completeness labels may be generated when the classifier misclassifies samples even on complete data. This is a fundamental limitation of pseudo-label-based approaches that do not rely on human annotations. To assess the prevalence of such failure cases, we evaluated the performance of the pre-trained classifier on each dataset. Table~\ref{tab:tpsc_classifier_accuracy} shows that MOSEI exhibits relatively lower accuracy compared to other datasets, likely due to its higher complexity. This suggests that the noisier pseudo-labels in MOSEI can partially explain its relatively weaker generalization performance.

We further analyzed failure cases and observed that misclassifications mainly occur in ambiguous and sarcastic expressions, as shown in Table~\ref{tab:tpsc_failure_cases}. These cases are inherently challenging to predict using only the text modality. Despite these limitations, the impact of such mispredictions on completeness estimation is limited. Due to their low TCP, TPSC assigns low completeness scores to misclassified samples, preventing overestimation of completeness and encouraging reliance on proxy features.

\section{Analyses with competitive baselines}
\label{sec:analyses_baseline}

\subsection{Metric-Level Comparison}
\label{sec:metric_level_comp}

In MSA tasks, both classification and regression performances are evaluated to provide a comprehensive assessment. However, classification metrics may not fully represent the quality of sentiment prediction in certain cases due to discretization in their computation. 

Specifically, this discretization process discards fine-grained differences between predictions within the same rounding interval. For example, when computing the classification metrics (e.g., Acc-7, Acc-5), continuous sentiment predictions are first converted into discrete classes via rounding, which may lead to inconsistencies near class boundaries, as follows:

{\normalsize
\begin{itemize}
\setlength{\itemsep}{2pt}
\setlength{\leftmargini}{1.2em}
\item GT: $1.4 \to \mathrm{round}(1.4) \to$ class~1
\item Model A: $1.55$ (MAE $=0.15$) $\to$ class~2
\item Model B: $1.10$ (MAE $=0.30$) $\to$ class~1
\end{itemize}
} 

As shown in Table~\ref{tab:tcmr_vs_cenet_prmf}, except for a few classification metrics, TCMR-TPSC consistently outperforms CENet and P-RMF across all benchmark datasets. This indicates that TCMR provides more reliable sentiment prediction overall.

\subsection{Training Efficiency Analysis}

\begin{table}[h!]
\centering
\small
\caption{Comparison of model complexity and training time per epoch.}
\label{tab:complexity}
\begin{tabular}{lcc}
\toprule
Model & \# Params (M) & Time / Epoch (s) \\
\midrule
LNLN      & 115.97 & 15.40 \\
P-RMF     & 117.31 & 18.00 \\
TF-Mamba  & 111.30 & 11.12 \\
TCMR      & 119.23 & 12.48 \\
\bottomrule
\end{tabular}
\end{table}

As shown in Table~\ref{tab:complexity}, TCMR achieves the second-fastest training speed, following TF-Mamba. This is because the reconstruction modules used in LNLN and P-RMF involve more computationally intensive operations, whereas TCMR employs a lightweight MLP-based CompNet. As a result, the additional optimization steps introduced by AOS do not significantly increase the overall training cost.

\begin{figure}[hbt]
    \centering
    \includegraphics[width=\linewidth]{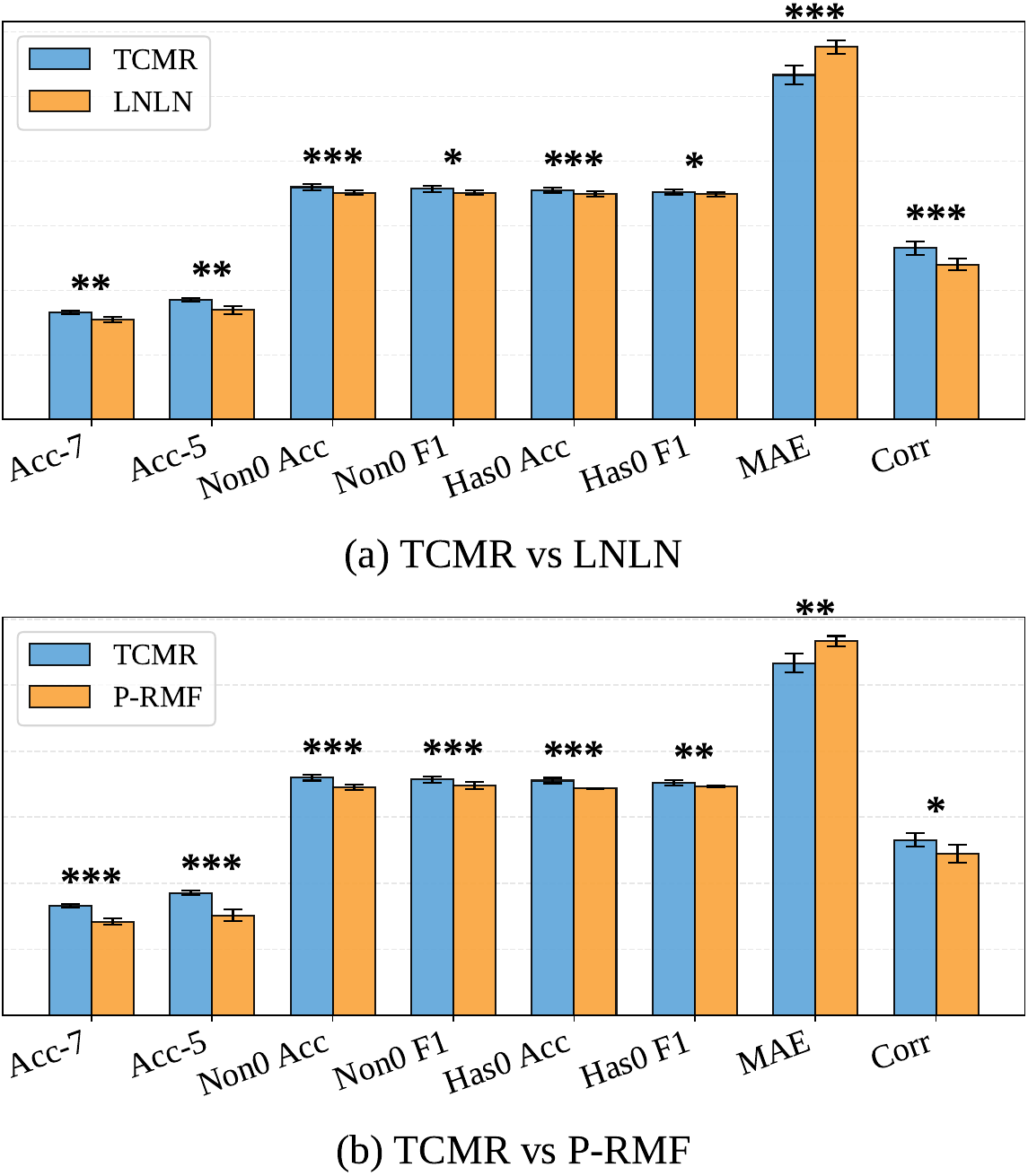}
    \caption{Error bars indicate standard deviations. Asterisks denote
statistical significance (* p < 0.1, ** p < 0.05, *** p < 0.01).}
    \label{fig:mean_std}
\end{figure}

\subsection{Statistical Analysis}

To examine the statistical significance of performance differences among models, we conduct comparative experiments on the MOSI dataset with LNLN, P-RMF, and TCMR. As shown in Figure~\ref{fig:mean_std}, TCMR achieves statistically significant performance improvements across most evaluation metrics.


\end{document}